\documentclass{article}

\usepackage{arxiv}

\usepackage[utf8]{inputenc} % allow utf-8 input
\usepackage[T1]{fontenc}    % use 8-bit T1 fonts
\usepackage{hyperref}       % hyperlinks
\usepackage{url}            % simple URL typesetting
\usepackage{booktabs}       % professional-quality tables
\usepackage{amsfonts}       % blackboard math symbols
\usepackage{nicefrac}       % compact symbols for 1/2, etc.
\usepackage{microtype}      % microtypography
\usepackage{lipsum}
\usepackage{graphicx}

\usepackage{color,soul}
\usepackage{makecell}
\usepackage{lineno,hyperref}
\usepackage{fancyvrb}
\usepackage{float}
\usepackage{amsmath}

\floatstyle{plain}
\newfloat{listing}{thp}{lop}
\floatname{listing}{Listing}

\title{Deliberative and Conceptual Inference in Service Robots}

\author{
 Luis A. Pineda \\
  Department of Computer Science\\
  Instituto de Investigaciones en Matem\'aticas Aplicadas y en Sistemas\\
  Universidad Nacional Aut\'onoma de M\'exico \\
  \texttt{lpineda@unam.mx} \\
   \And
 No\'e Hern\'andez \\
  Department of Computer Science\\
  Instituto de Investigaciones en Matem\'aticas Aplicadas y en Sistemas\\
  Universidad Nacional Aut\'onoma de M\'exico \\
  \texttt{nohernan@turing.iimas.unam.mx} \\
  \And
 Arturo Rodr\'iguez \\
  Facultad de Estudios Superiores Arag\'on\\
  Universidad Nacional Aut\'onoma de M\'exico\\
  Pittsburgh, PA 15213 \\
  \texttt{arturorodriguez35@aragon.unam.mx} \\
  \And
 Ricardo Cruz\\
  Department of Computer Science\\
  Instituto de Investigaciones en Matem\'aticas Aplicadas y en Sistemas\\
  Universidad Nacional Aut\'onoma de M\'exico \\
  \texttt{cricardo.cruz@iimas.unam.mx} \\
  \And
 Gibrán Fuentes\\
  Department of Computer Science\\
  Instituto de Investigaciones en Matem\'aticas Aplicadas y en Sistemas\\
  Universidad Nacional Aut\'onoma de M\'exico \\
  \texttt{gibranfp@unam.mx} \\
}

\begin{document}
\maketitle
\begin{abstract}
Service robots need to reason to support people in daily life situations. Reasoning is an expensive resource that should be used on demand whenever the expectations of the robot do not match the situation of the world and the execution of the task is broken down; in such scenarios the robot must perform \emph{the common sense daily life inference cycle} consisting on diagnosing what happened, deciding what to do about it, and inducing and executing a plan, recurring in such behavior until the service task can be resumed. Here we examine two strategies to implement this cycle: (1) a pipe-line strategy involving abduction, decision-making and planning, which we call \emph{deliberative inference} and (2) the use of the knowledge and preferences stored in the robot's knowledge-base, which we call \emph{conceptual inference}. The former involves an explicit definition of a problem-space that is explored through heuristic search, and the latter is based on conceptual knowledge including the human user preferences, and its representation requires a non-monotonic knowledge-based system. We compare the strengths and limitations of both approaches. We also describe a service robot conceptual model and architecture capable of supporting the daily life inference cycle during the execution of a robotics service task. The model is centered in the declarative specification and interpretation of robot's communication and task structure. We also show the implementation of this framework in the fully autonomous robot Golem-III. The framework is illustrated with two demonstration scenarios. 
\end{abstract}

% keywords can be removed

\section{Inference in Service Robots}
\label{sec:introduction}

\begin{figure}
\includegraphics[width=1.0\textwidth]{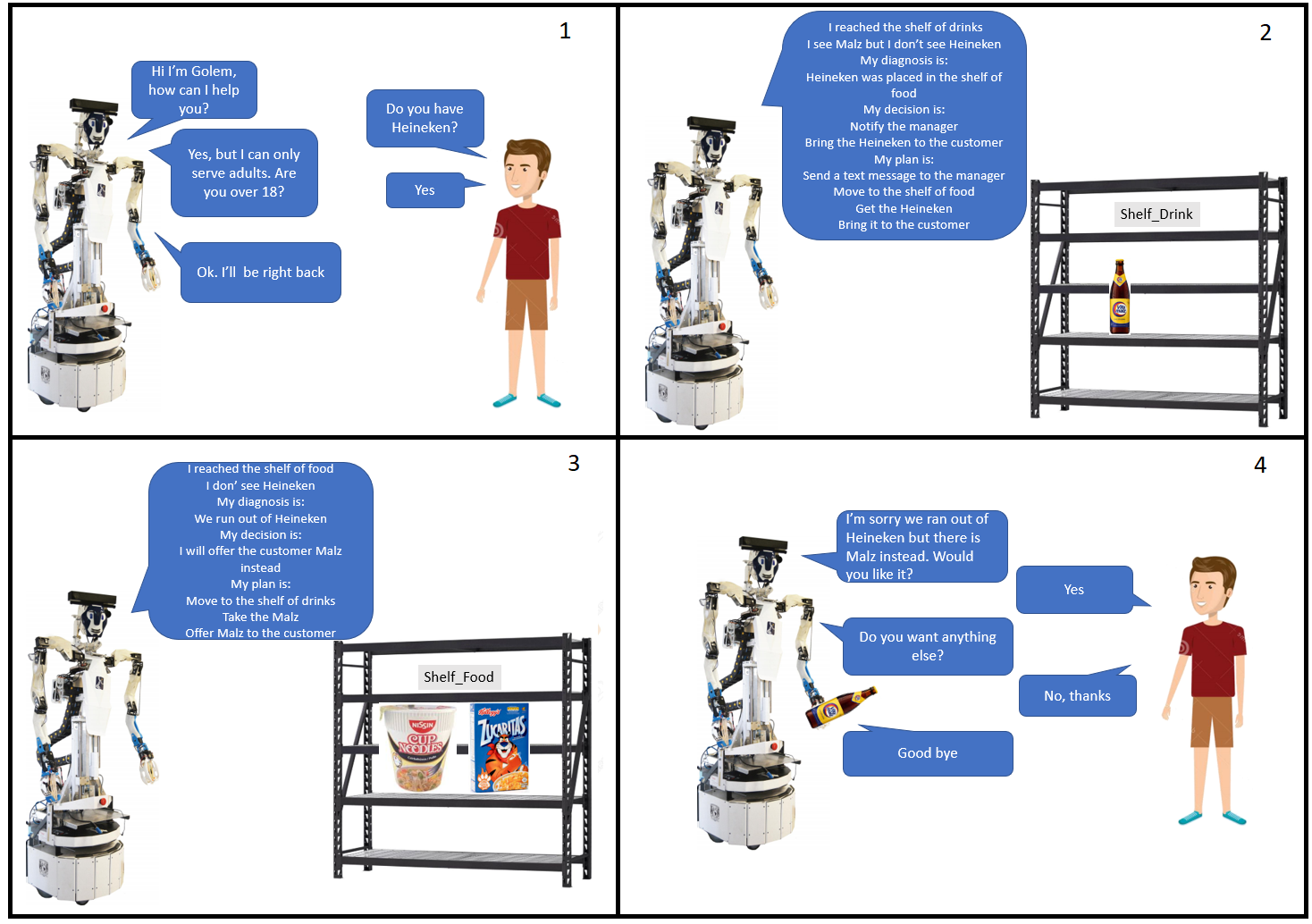}
\centering
\caption{Common Sense Daily-Life Inference Cycle.}
\label{fig:supermarket-scenario}
\end{figure}

Fully autonomous service robots aimed to support people in common daily tasks require competence in an ample range of faculties, such as perception, language, thought and motor behavior, whose deployment should be highly coordinated for the execution of service robotics tasks. A hypothetical illustrative scenario in which a general purpose service robot performs as a supermarket assistant is shown in the story-board in Figure \ref{fig:supermarket-scenario}. The overall purposes of the robot in the present scenario are i) to attend the customer's information and action requests or commands; ii) to keep the supermarket in order, and iii) to keep the manager informed about the state of the inventory in the stands. The basic behavior can be specified schematically but if the world varies in unexpected ways, due to spontaneous behavior of other agents or to unexpected natural events, the robot has to reason to complete the service tasks successfully. In box (1) the robot greets the customer and offers help, and the customer asks for a beer. The command is performed as an indirect speech act in the form of a question. The robot has conceptual knowledge stored in his knowledge-base including the obligations and preferences of the agents involved. In this case, the restriction that alcoholic beverages can only be served to people over eighteen. This prompts the robot to issue an information request to confirm the age of the customer. When the user does so, the robot is ready to accomplish the task. The robot has a scheme to deliver the order and also knowledge about the kinds of objects in the supermarket, including their locations. So, if every thing is as expected the robot can accomplish the task successfully by executing the scheme. With this information the robot moves to the stand of drinks where the beer should be found. However, in the present scenario the Heineken is not there, the scheme is broken down and to proceed the robot needs to reason. As this is an expensive resource it should be employed on demand. 

The reasoning process involves three main kinds of inference: 

\begin{enumerate}
\item 
An abductive inference process to the effect of diagnosing the cause of the current state of the world which differs from the expected one. 
\item 
A decision making inference to the effect of deciding what to do to produce the desired state on the basis of the diagnosis and the overall purposes of the agent in the task.
\item
The decision made becomes the goal of a plan that has to be induced and carried out to produce the desired state of the world.
\end{enumerate}

We refer to this cycle as the \emph{common sense daily-life inference cycle}. It also has to be considered that the world may have changed along the development of the task or that the robot may fail in achieving some actions of the plan, and it needs to check again along the way; if the world is as expected the execution of the plan is continued but if something is wrong, the robot needs to engage recurrently in the daily-life inference cycle until the task is completed or the robot needs to give up.

The first instance of the daily-life inference cycle in the scenario in Figure \ref{fig:supermarket-scenario} is shown in box (2). The inference is prompted by the robot's visual behavior, which is goal directed, that fails to recognize the intended object. This failure is reported through the declarative speech act \emph{I see Malz but I don't see the Heineken}. Then, the robot performs a diagnosis inference to the effect of determining what went wrong. This kind of reasoning proceeds from an observation to its causes, has an abductive character and is non-monotonic. In this case the robot hypothesizes where the beer should be and what was the cause of such state (i.e., \emph{Heineken was placed on the shelf of food}). The decision about what to do next involves two goals: informing the manager the state of the inventory of the drinks stand through a text message --not illustrated in the figure-- and delivering the beer to the customer, and a plan to such an effect is induced and executed.

The robot carries on with the plan but fails to find the beer in the stand for food and the daily-life inference cycle is invoked again, as shown in box (3). The diagnosis this time is that the supermarket ran out of beer and the decision is to offer the customer the Malz instead. The plan consists on moving back to the shelf of drinks, get the Malz (this action is not shown in the figure), make the offer and conclude the task, as illustrated in box (4).

The implementation of the scenario relies on two conceptual engines that work together. The first is a methodology and programming environment for specifying and interpreting conversational protocols, that here we call \emph{dialogue models}, that carry on with the schematic behavior, by issuing and interpreting the relevant speech acts during the execution of the task. Dialogue models have two dual aspects and represent both the task structure and the communication structure, which proceed in tandem. The second consists on the inferential machinery that is used on demand, which is called upon within the interpretation of the input and output speech acts. In this paper, we show the conceptual model and physical machinery to support the conversational protocols and inference capabilities require to achieve the kind of service tasks illustrated in Figure \ref{fig:supermarket-scenario}.

The structure of this paper is as follows: A summary of the relevant work on deliberative and conceptual inference in service robots is presented in Section \ref{sec:related-work}. Next, in Section \ref{sec:model}, we describe the conceptual model and architecture required to support the inferences deployed by the robot in the present scenario. In Section \ref{sec:sitlog} the $SitLog$ programming language for the specification dialogue models or interaction protocols in which inference is used on demand is reviewed. $SitLog$ supports the definition of robotics tasks and behaviors, which is the subject of Section \ref{sec:tasks-behaviors}. The non-monotonic service used to perform conceptual inferences is reviewed in Section \ref{sec:kb-system}. With this machinery in hand we present two strategies to implement the daily-life inference cycle. First in Section \ref{sec:deliberative-inference} we show the pipe-line strategy involving the definition of an explicit problem space and heuristic search. We describe a full demonstration scenario in which the robot Golem-III performs as a supermarket assistant. This demo was performed successfully at the final of the RoboCup German Open 2018 in the category @Home. Then in Section \ref{sec:conceptual-inference} we describe a second scenario in which Golem-III performs as a home assistant; here the daily-life inference cycle is performed on the bases of the knowledge of the scenario and the preferences of the user stored in the knowledge-base. Finally, in Section \ref{sec:conclusions} we discuss the advantage and limitations of both approaches and suggest further work to model better the common sense inferences made by people in practical tasks. 

\section{Related Work}
\label{sec:related-work}
Service robotics research has traditionally focused on tackling specific functionalities or carrying out tasks that integrate such functionalities (e.g. navigation~\cite{Durham2008,Grisetti2007}, manipulation~\cite{trex, herb2.0} or vision~\cite{moped, Espinace2013}). However, relatively few efforts have been made to define an articulated concept of service robots with general programming and inference architectures to develop high-level tasks. In this section, we briefly review related works on high-level programming languages and knowledge representation and reasoning systems for service robots. 

High-level task programming has been widely studied in robotics and several domain-specific languages and specialized libraries as well as extensions to general-purpose programming languages have been developed for this purpose. Many of these approaches are built upon finite state machines and extensions~\cite{xabsl, xrobots, smach}, although situation calculus~\cite{readylog, schiffer_golog, golex} and other formalisms are also common. Notable instances of domain-specific languages are the Task Description Language~\cite{tdl}, the Extensible Agent Behaviour Specification Language (XABSL)~\cite{xabsl}, XRobots~\cite{xrobots} and \texttt{ReadyLog}~\cite{readylog}. More recently, specialized frameworks and libraries, such as TREX~\cite{trex} and SMACH~\cite{smach}, have become attractive alternatives for high-level task programming. 

Reasoning is an essential ability in order for service robots to autonomously operate in a realistic scenario, robustly handling its inherent uncertainty and complexity. Existing reasoning systems are often employed for task planing, sometimes taking into account spatial and temporal information (e.g.~\cite{galindo2008, Lim2011,Karg2012}). These systems typically exploit logical~\cite{Schiffer2012} or probabilistic inference (e.g. Partially Observable Markov Decision Processes and variants), or combinations of both~\cite{pomdp_nav, pomdp_manipulation, Zhang2012a}. 

Reasoning systems rely on knowledge-bases to store and retrieve knowledge about the world, which can be obtained beforehand or dynamically by interacting with the users and the environment. One of the most prominent systems is the Knowledge processing for robots (KnowRob)~\cite{tenorth2013knowrob, Tenorth2017}, which has multiple knowledge representation and reasoning capabilities, and has been deployed in several complex tasks~\cite{Pangercic, becker2011pr2, fan2014, knowrob-map}.
Non-monotonic knowledge representation and reasoning systems are typically based on Answer Set Programming (ASP) (e.g.~\cite{Zhang2012a, Chen2010, opfer_asp}), some of which have been demonstrated in different complex tasks~\cite{jakob_asp, Chen2010, Chen2013, Chen2014, Chen2015}.

In this work, we present a general framework for deliberative and conceptual inference in service robots that its integrated within an interaction-oriented cognitive architecture and accessed through a domain-specific task programming language. This framework implements a light and dynamic knowledge-base system that allows non-monotonic reasoning and the specification of preferences.

\section{Conceptual Model and Robotics Architecture}
\label{sec:model}

To address the complexity described in Section \ref{sec:introduction} we have developed an overall conceptual model of service robots  \cite{service-robot-2014}  and the Interaction-Oriented Cognitive Architecture (IOCA) \cite{ioca-2011} for its implementation. Conceptual and deliberative inferences are the highest functions performed by the robot and are placed at the top of such functionality.

The conceptual model  is inspired in Marr's hierarchy of systems levels \cite{Marr:1982:VCI:1095712} and consists of the functional, the algorithmic and the implementation system levels. The functional level is related to \emph{what} the robot does from the point of view of the human-user and focuses on the definition of tools and methodologies for the declarative specification and interpretation of robotic tasks and behaviors; the algorithmic level consists on the specification of \emph{how} behaviors are performed and focuses on the the development of robotics algorithms and devices; finally, the implementation level focuses on system programming, operating systems and the software agent's specification and coordination.

The center of the conceptual model is the interpreter of SitLog \cite{sitlog-2013}. This is a programming language for the declarative specification and interpretation of the robot's communication and task structure. SitLog defines two main abstract data-types: the \emph{situation} and the \emph{dialogue model} ({DM}). A situation is an information state defined in terms of the expectations of the robot and a DM is a directed graph of situations representing the task structure. Situations can be \emph{grounded} and correspond to an actual spatial and temporal estate of the robot, where concrete perceptual and motor actions are performed, or \emph{recursive}, consisting on a full dialogue model, possibly including recursive situations, permitting the definition of large abstractions of the task structure.

Dialogue models have a dual interpretation as conversational or speech acts protocols that the robot performs along the execution of a task. From this perspective, expectations are the speech acts that can potentially be expressed by an interlocutor, either human or machine, in the current situation. Actions are thought of as the speech acts performed by the robot as a response to such interpretations. Events in the world that can occur in the situation are also considered expectations that give rise to intentional action by the robot. For this reason dialogue models represent the communication or interaction structure, and correspond to the task structure.

\begin{figure}
\includegraphics[width=0.5\textwidth]{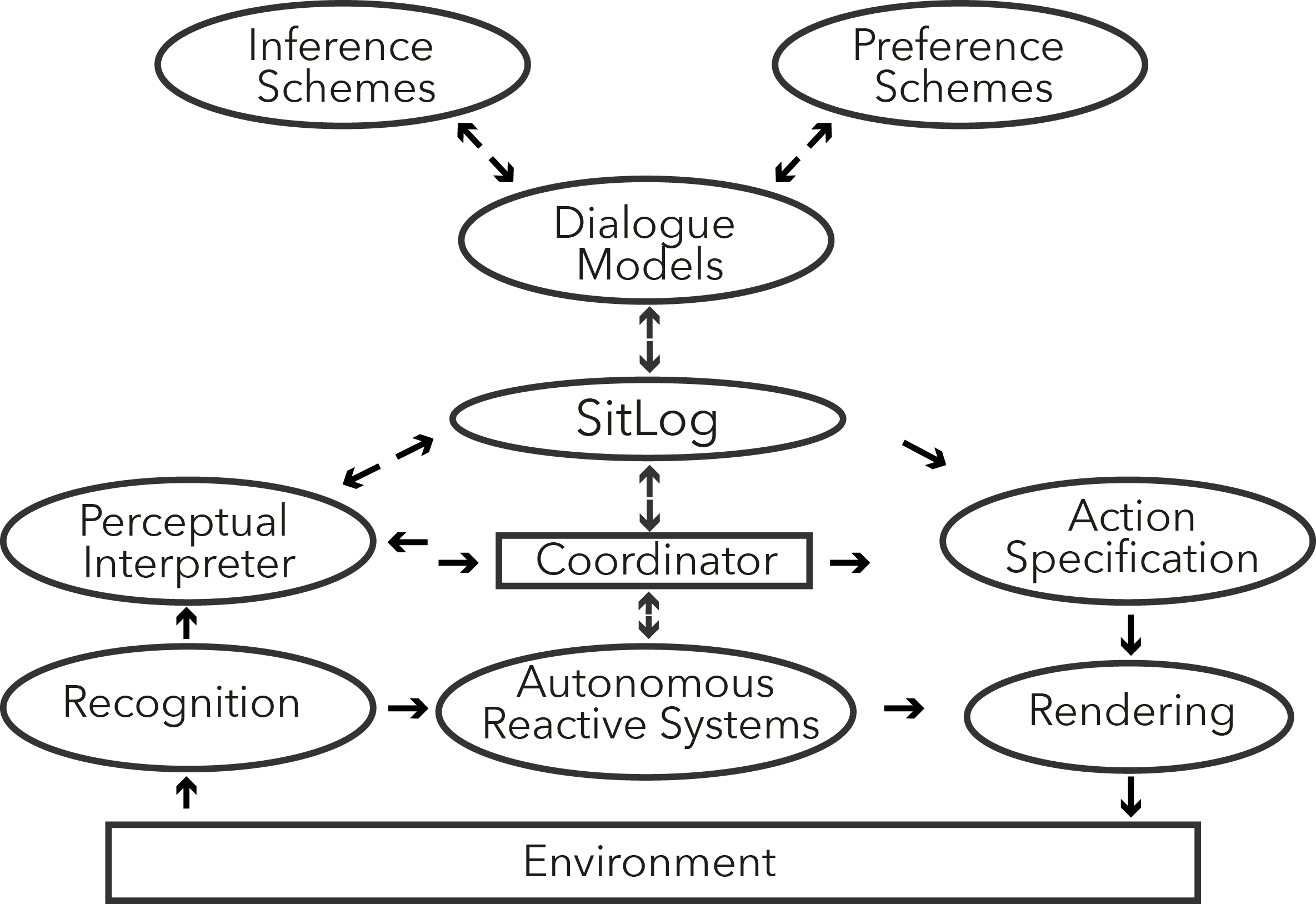}
\centering
\caption{Interaction-Oriented Cognitive Architecture (IOCA).}
\label{ioca}
\end{figure}

The Interaction-Oriented Cognitive Architecture is illustrated in Figure \ref{ioca}. The architecture includes a low level reactive cycle involving low-level recognition and rendering of behaviors that are managed by the Autonomous Reactive System directly. This cycle is embedded within the communication or interaction cycle and has the SitLog's interpreter as its center, which performs the interpretation of the input and the specification of the output in relation to current situation and dialogue model. The reactive and communication cycles normally proceed independently and continuously, the former working at as least one order of magnitude faster than the latter, although there are times in which one needs to take full control of the task for performing a particular process, and there is a loose coordination between the two cycles.

The perceptual interpreters are modality specific and receive the output of the low-level recognition process bottom-up but also the expectations in the current situation top-down, narrowing the possible interpretations of the input. There is one perceptual interpreter for each input modality which instantiates the expectation that is meaningful in relation to the context. Expectations are hence \emph{representations of interpretations}. The perceptual interpreters promote sub-symbolic information produced by the input modalities into a fully articulated representation of the world, as seen by the robot in the situation. Standard perception and action robotics algorithms are embedded within modality specific perceptual interpreters for the input and for specifying the external output respectively.

Knowledge and inference resources are used on demand within the conversational context. These ``thinking" resources are also actions, but unlike perceptual and motor actions, that are directed to the interaction with the world, thinking is an internal action that mediates the input and output, permitting the robot to anticipate and cope better with the external world. The communication and interaction cycle is then the center of conceptual architecture and is oriented to interpret and act upon the world, but also to manage thinking resources, that are embedded within the interaction, and the interpreter of SitLog coordinates the overall intentional behavior of the robot.

The present conceptual architecture supports rich and varied but schematic or stereotyped behavior. Task structure can proceed as long as at least one expectation in the current situation is met by the robot. However, schematic behavior can easily break down in dynamic worlds when none or more than one expectations are satisfied in the situation. When this happens the interpretation context is lost and the robot needs to recover it to continue with the task. There are two strategies to deal with such contingencies: 1) to invoke domain independent dialogue models for task management, which here we refer to as \emph{recovery protocols} or 2) to invoke inference strategies to recover the ground. In this latter case the robot requires to make an abductive inference or a diagnosis in order to find out why none of its expectations was meet, decide the action needed to recover the ground on the basis of such diagnosis, in conjunction with a given set of preferences or obligations, and induce and execute a plan to achieve such goal. Here we refer to the cycle of diagnosis-decision making-planning as the \emph{daily life inference cycle} which is specified in Section \ref{sec:daily-life-inf-cycle}. This cycle is invoked by the robot when schematic behavior cannot proceed and a recovery protocol that is likely to recover the ground is not available.

\section{The SitLog Programming Language}
\label{sec:sitlog}

The overall intelligent behavior of the robot in the present framework depends on the synergy of intentional dialogues oriented to achieve the goals of the task and the inference resources that are used on demand within such purposeful interaction. The code implementing the \emph{SitLog} programming language is available as a GitHub repository at \url{https://github.com/SitLog/source_code/}. 

% In this section we review the \emph{SitLog} programming language to specify such intentional interaction \cite{sitlog-2013}; next, in Section \ref{sec:tasks-behaviors}, we show how \emph{SitLog} is used to specify robotics tasks and behaviors and in Sections \ref{sec:deliberative-inference} and \ref{sec:conceptual-inference} the deliberative and conceptual inference resources are discussed respectively.

\subsection{\emph{SitLog}'s basic abstract data-types}
\label{sec:sitlog-data-types}

The basic notion of \emph{SitLog} is the \emph{situation}. A situation is an information state characterized by the expectations of an intentional agent, such that the agent --the robot-- remains in the situation as long as its expectation are the same. This notion provides a large spatial and temporal abstraction of the information state because although there may be large changes in the world or in the knowledge that the agent has in the situation, its expectations may nevertheless remain the same.

A situation is specified as a set of expectations. Each expectation shas an associated action that is performed by the agent when such expectation is meet in the world, and the situation that is reached as a result of such action. If the set of expectations of the robot after performing such action remain the same, the robot recurs to the same situation. Situations, actions and next situations may be specified concretely but \emph{SitLog} also allows that these knowledge objects are specified through functions, possible higher-order, that are evaluated in relation to the interpretation context. The results of such evaluation are concrete interpretations and actions that are performed by the robot, as well as the concrete situation that is reached next in the robotics task. Hence, a basic task can be modeled with a directed graph with a moderate and normally small set of situations. Such directed graph is referred to in \emph{SitLog} as a \emph{Dialogue Model}. Dialogue models can have recursive situations including a full dialogue model, providing the means for expressing large abstractions and modeling complex tasks. A situation is specified in \emph{SitLog} as an attribute-value structure, as follows:

\begin{verbatim}[fontfamily=courier]
[
  id ==> Situation_ID(Arg_List),
  type ==> Situtation_Type_ID,
  in_arg ==> In_Arg,
  out_arg ==> Out_Arg,
  prog ==> Expression,
  arcs ==> [
             Expect1:Action1 => Next_Sit1,
             Expect2:Action2 => Next_Sit2,
 		     ...
             Expectn:Actionn => Next_Sitn
           ]
]
\end{verbatim}

\emph{SitLog}'s interpreter is programmed fully in Prolog and subsumes Prolog's notation. Following Prolog's standard conventions strings starting with lower and upper case letters are \emph{atoms} and \emph{variables} respectively, and the \texttt{==>} is an operator relating an attribute with its corresponding value. Values are expressions of a functional language including constants, variables, operators, predicates, and operators such as unification, variable assignment and the \emph{apply} operator for dynamic binding and evaluation of functions. The functional language supports the expression of higher-order functions too. The interpretation of a situation by \emph{SitLog} consists on the interpretation of all its attributes from top to bottom. The attributes \texttt{id}, \texttt{type} and \texttt{arcs} are mandatory. The value of \texttt{prog} is a list of expressions of the functional language, and in case such attribute is defined it is evaluated unconditionally before the \texttt{arcs} attribute.

A dialogue model is defined as a set of situations. Each DM has a designated initial situation and at least a final one. A \emph{SitLog} program consists on a set of DMs, one designed as the \emph{main} DM. This may include a number of situations of type \texttt{recursive} each containing a full DM. \emph{SitLog}'s interpreter unfolds a concrete graph of situations, starting form the initial situation of the main DM and generates a Recursive Transition Network (RTN) of concrete situations. Thus, the basic expressive power of \emph{SitLog}'s corresponds to a context-free grammar. \emph{SitLog}'s interpreter consists of the coordinated tandem operation of the RTN's interpreter that unfolds the graph of situations and the functional language that evaluates the attributes' values.

All links of the \texttt{arcs} attribute are interpreted during the interpretation of a situation. Expectations are sent to the perceptual interpreter top-down, which instantiates the expectation that is met by the information provided by the low-level recognition processes, and sends such expectation back --bottom-up-- to \emph{SitLog}'s interpreter. From the perspective of the declarative specification of the task, \texttt{EXPECTn} contains the information provided by perception. Once an expectation is selected the corresponding action and next situation are processed. In this way, \emph{SitLog} abstracts over the external input, and such interface is transparent for the user in the declarative specification of the task.  Expectations and actions may be empty, in which case a transition between situations is performed unconditionally.

\subsection{\emph{SitLog}'s programming environment}
\label{sec:sitlog-prog-env}

\emph{SitLog}'s programming environment includes the specification of a set of global variables that have scope over the whole of the program, and a set of local variables that have scope over the situations of a particular dialogue model. \emph{SitLog}'s variables are also defined as attribute-value pairs, the attribute being a Prolog's atom and its value a standard Prolog's variable. \emph{SitLog} variables can have arbitrary Prolog's expressions as their values. All variables can have default values and can be updated through the standard assignment operator, which is defined in \emph{SitLog}'s functional language. 

Dialogue models and situations can also have arguments, whose values are handled by reference in the programming environment, and dialogue models and situations allow input and output values that can propagate through the graph by these interface means. 

The programming environment includes also a \emph{pipe} global communication structure that provides an input to the main DM and propagates through all concrete DMs and situations that unfold in the execution of the task. This channel is stated through the attributes \texttt{in\_arg} and \texttt{out\_arg}, whose definition is optional. The value of \texttt{out\_arg} is not specified when the situation is called upon --i.e., it is a variable--  and can be given a value in the body of the situation through a variable assignment or through unification. In case there is no such assignment, the input and output pipes are unified when the interpretation of the situation is concluded. The value of \texttt{in\_arg} can be also underspecified, and given a value within the body of the situation too. In case these are not stated explicitly the value of \texttt{in\_arg} propagates to \texttt{out\_arg} by default as was mentioned. 

Global and local variables, as well as the values of the pipe, have scope over the local program and within each link of the \texttt{arcs} attribute, and their values can be changed through the \emph{SitLog}'s assignment operator or through unification. However, a local program and links are encapsulated local objects that have no scope outside their particular definition. Hence, Prolog's variables defined in \texttt{prog} and in different links of \texttt{arcs} attribute are not bounded even if they have the same name. The strict locality of these programs has proved to be very effective for the definition of complex applications.

The programming environment includes as well a \emph{history} of the task conformed by the stack structure of all dialogue models and situations, with their corresponding concrete expectation, action and next situation, unfolded along the execution of the task. The current history can be consulted through a function of the functional language, and can be used not only to report what happened before but also to make a decision about the future course of action. 

The elements of the programming environment augment the expressive power of \emph{SitLog}, which corresponds overall to a context-sensitive grammar.  The representation of the task is hence very expressive but preserves still the graph structure, and \emph{SitLog} provides a very good compromise between expressiveness and computational cost.

\subsection{\emph{SitLog}'s diagrammatic representation}
\label{sec:sitlog-diag-rep}

\emph{SitLog} programs have a diagrammatic representation as illustrated in Figure \ref{fig:dummy}.\footnote{The full commented code of the present \emph{SitLog} program is given in Appendix \emph{An axample program in SitLog}.} DMs are bounded by large dotted ovals including the corresponding situation's graph (i.e., \emph{main} and \emph{wait}). Situation are represented by circles with labels indicating the situation's id and type. In the example the \emph{main} DM has three situations whose id's are \texttt{is}, \texttt{fs} and \texttt{rs}. The situation identifier is optional --provided by the user--  except the initial situation that has the mandatory id \texttt{is}. The types id's are also optional with the exception of the \texttt{final} and \texttt{recursive}, as these are used by \emph{SitLog} to control the stack of DMs. The links between situations are labeled with pairs of form $\alpha$:$\beta$, which stand for expectations and actions respectively. When the next situation is specified concretely the arrow linking to situations is stated directly; however, if the next situation is stated through a function (e.g. \emph{h}), there is a large bold dot after the $\alpha$:$\beta$ pair with two or more exit arrows. This indicates that the next situation depends on the value of \emph{h} in relation to the current context, and there is a particular next situation for each particular value. For instance, the edge of \texttt{is} in de DM \emph{main} that is labeled by $[\text{\tt day},f]$:$g$, representing the expectation as a list of the value of the local variable \texttt{day} and the function $f$ and the action as the function $g$, is followed by a large black dot labeled by the function \emph{h}; this function has two possible values, one cycling back into \texttt{is} and the other leading to the recursive situation \texttt{rs}. This depicts when the expectation that is met at the initial situation satisfies the value of the local variable \texttt{day} and the value of function \emph{f}, so the action defined by the value of function \emph{g} is performed, and the next situation is the value of function \emph{h}.

The circles representing recursive situations have also large internal dots representing control return points from embedded dialogue models. The dots mark the origin of the exit links that have to be transversal  whenever the execution of an embedded DM is concluded, when the embedding DM is popped up from the stack and resumes execution. The labels of the expectations of such arcs and the names of the corresponding final states of the embedded DM are the same, depicting that the expectations of a recursive situation correspond to the designated final states of the embedded DM. This is the only case in which an expectation is made available internally to the interpreter of \emph{SitLog} and is not provided by a perceptual interpreter as a result of an observation from the external world.

Finally, the bold arrows depict the information flow between dialogue models. The output bold arrow leaving \emph{main} at the upper right corner denotes the value of \texttt{out\_arg} when the task is concluded. The bold arrow from \emph{main} to \emph{wait} denotes the corresponding pipe connection, such that the value of \texttt{out\_arg} of the situation \texttt{rs} in  \emph{main} is the same as the value of \texttt{in\_arg} in the initial situation of \emph{wait}. The diagram also illustrates that the value of  \texttt{in\_arg} propagates back to \emph{main} through the value of \texttt{out\_arg} in both  final situations \texttt{fs1} and \texttt{fs2}; since the attribute \texttt{out\_arg} is never set within the DM \emph{wait}, the original value of \texttt{in\_arg} keeps passing on through all the situations including the final ones. The expectations of the arcs of \texttt{is} in the DM \emph{wait} take the input from the perceptual interpreter being either the value of \texttt{in\_arg} or the atom \texttt{loop}.

\begin{figure}[tb]
\includegraphics[width=0.6\textwidth]{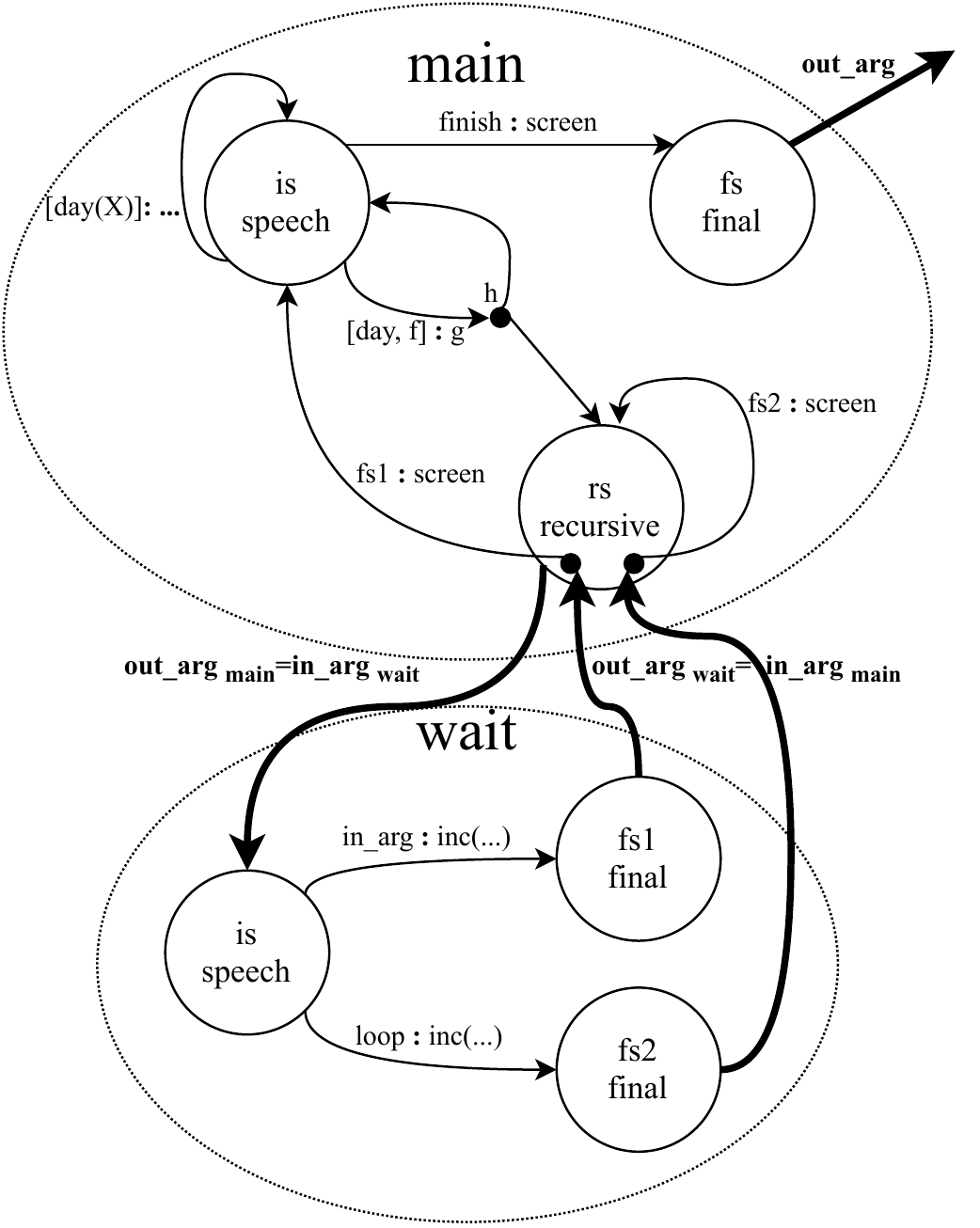}
\centering
\caption{Graphical representation of an example dialogue model written in \emph{SitLog}.}
\label{fig:dummy}
\end{figure}

\section{Specification of Task Structure and Robotics Behaviors}
\label{sec:tasks-behaviors}

The functional system level addresses the tools and methodologies to define the robot's competence. In the present model such competence depends on a set of robotics behaviors and a composition mechanism to specify complex tasks. Behaviors rely on a set of primitive perceptual and motor actions. There is a library of such basic actions, each associated to a particular robotics algorithm. Such algorithms constitute the ``innate" capabilities of the robot. 

In the present framework, robotics behaviors are $SitLog$ programs whose purpose is to achieve a specific goal by executing one or more basic actions within a behavior's specific logic. Examples of such behaviors are \emph{move}, \emph{see}, \emph{see\_object}, \emph{approach}, \emph{take}, \emph{grasp}, \emph{deliver}, \emph{relieve}, \emph{see\_person}, \emph{detect\_face},  \emph{memorize\_face}, \emph{reconize\_face}, \emph{point}, \emph{follow}, \emph{guide}, \emph{say}, \emph{ask}, etc. The $SitLog$'s code of \emph{grasp}, for instance, is available at the GitHub repository of $SitLog$ \url{https://bit.ly/grasp-dm}.

Behaviors are parametric abstract units that can be used as atomic objects but can also be defined as structured objects using other behaviors. For instance, \emph{take} is a composite behavior using \emph{approach} and \emph{grasp} and \emph{deliver} uses \emph{move} and \emph{relieve}. Another example is \emph{see} that uses \emph{see\_object}, \emph{see\_person} and \emph{see\_gesture} to interpret a visual scene generally.

All behaviors have a number of terminating \emph{status}. If the behavior is executed successfully the status is \emph{ok}; however, there may be a number of failure conditions, particular to the behavior, that may prevent its correct termination, and each is associated with a particular \emph{error} status. The dialogue model at the application layer should consider all possible status of all behaviors in order to improve the robot's reliability.

Through these mechanisms complex behaviors can be defined, such as \emph{find}, that given a search path and a target object or person, enables the robot to explore the space using the \emph{scan} and \emph{tilt} behaviors to move its head and make visual observations at different positions and orientations. The full $SitLog$'s code of \emph{find} is also provided at the GitHub repository of $SitLog$ \url{https://bit.ly/find-dm}.

Behaviors should be quite general, robust and flexible, so they can be used in different tasks and domains. There is a library of behaviors that provide the basic capabilities of the robot from the perspective of the human-user. This library evolves with practice and experience and constitutes a rich empirical resource for the construction and application of service robots  \cite{service-robot-2014}.

The composition mechanism is provided by $SitLog$ too, which allows the specification of dialogue models that represent the tasks and communication structure. Situations in these latter dialogue models represent stages of a task, that can be partitioned into sub-tasks. So the tasks as a whole can be seen as a story-board, where each situation corresponds to a picture.

\begin{figure}
\includegraphics[width=0.6\textwidth]{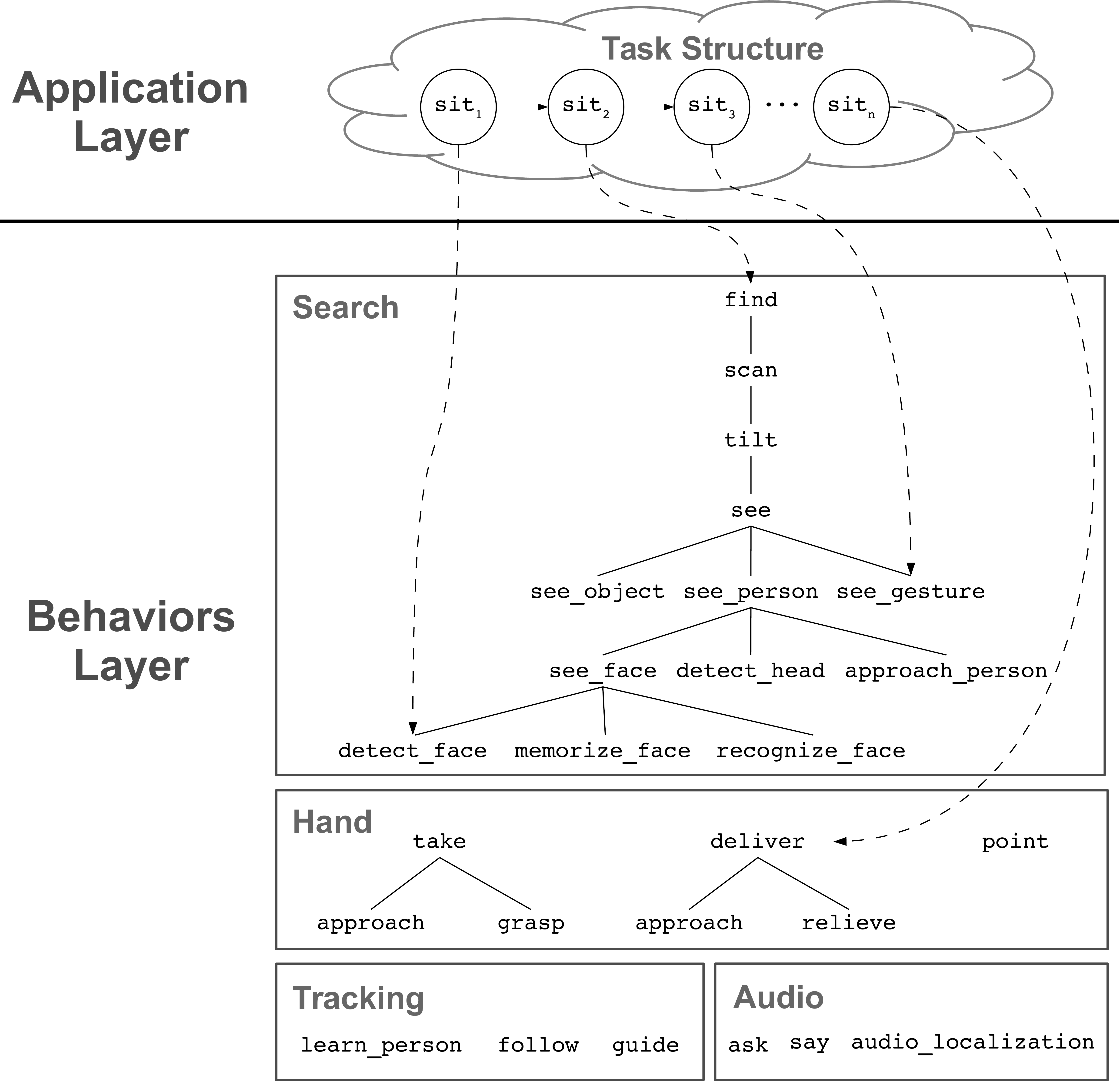}
\centering
\caption{Tasks and behaviors.}
\label{fig:task_behaviors}
\end{figure}

For example, if the robot performs as a supermarket assistant, the structure of the tasks can be construed as 1) take an order from the human customer; 2) find and take the requested product; and 3) deliver the product to the customer. These tasks correspond to the situations in the application layer, as illustrated in Figure \ref{fig:task_behaviors}. Situations can be further refined in several sub-tasks specified as more specific dialogue models embedded on the situations of upper levels, and the formalism can be used to model complex task quite effectively.

The dotted lines from the application to the behaviors layer in Figure \ref{fig:task_behaviors} illustrate that behaviors are used at the application layer as abstract units at different degrees of granularity. For instance, \emph{find} is used as an atomic behavior but also \emph{detect\_face} can be used directly by a situations at the level of the task structure, despite that \emph{detect\_face} is used by \emph{find}. The task structure at the application layer can be partitioned in subordinated tasks too. For this, $SitLog$'s supports the recursive specification of dialogue models and situations, enhancing the expressive power of the formalism.

Although both task structure and behaviors are specified through $SitLog$'s programs these correspond to two different levels of abstraction. The top level specifies the final application task-structure, and is defined by the final user, while the lower level consists on the library of behaviors, which should be generic and potentially useful in diverse application domains. 

From the point of view of an ideal division of labor, the behaviors layer is the responsibility of the robot's developing team, while the application's layer is the focus of teams oriented to the development of final service robot applications.

\subsection{The General Purpose Service Robot}

Prototypical or schematic robotic tasks can be defined through dialogue models directly. However, the structure of the task has to be known in advance, and there are many scenarios in which this information is not available. For this, in the present framework we define a general purpose mechanism that translates speech acts performed by the human-user into a sequence of behaviors, which is interpreted by a behavior's dispatcher one behavior at a time, and finishes the task when the list has been emptied \cite{service-robot-2014}. We refer to this mechanism as \emph{General Purpose Service Robot} or simply \emph{GPSR}.

In the basic case, all the behaviors in the list terminate with the status \emph{ok}. However, whenever the behaviors terminate with a different status, something in the world was unexpected, or the robot failed, and the dispatcher must take an appropriate action. We consider two main types of error situations. The first may be a general but common and known failure, in which case, a recovery protocol is invoked; these protocols are implemented as standard $SitLog$'s dialogue models, and undergo a procedure that is specific to fix the error, and when this is accomplished, they return control to the dispatcher, and continue with the task. The second type is about errors cannot be prevented, to recover from them the robot needs to engage in the daily-life inference cycle, as discussed in the Section \ref{sec:introduction}, and will be elaborated in Sections \ref{sec:daily-life-inf-cycle}, \ref{sec:deliberative-inference} and \ref{sec:conceptual-inference}.

\section{Non-Monotonic Knowledge-Base Service}
\label{sec:kb-system}

The specification of service robot's tasks requires an expressive, robust but flexible knowledge-base service. The robot may require to represent, reason and maintain terminological or linguistic knowledge, general and particular concepts about the world, and about the application domain. There may be also defaults, exceptions and preferences, which can be acquired and updated incrementally during the specification and execution of a task, and a non-monotonic KB service is required. Inferences about this kind of knowledge are referred to here as \emph{conceptual inferences}. 

To support such functionality we have developed a non-monotonic knowledge-base service based on the specification of a class hierarchy. This system supports the representation of  classes and individuals, which can have general or specific properties and relations \cite{Non-Monotonic-KB-2017, Preferences-SR-2018}. Classes and individuals are the primitive objects and constitute the ontology. There is a main or top class which includes all the individuals in the universe of discourse; this can be divided in a finite number of mutually exclusive partitions, each corresponding to a subordinated or subsumed class. Subordinated classes can be further partitioned into subordinated mutually exclusive partitions giving rise to a strict hierarchy of an arbitrary depth, and classes are related through a proper inclusion relation. Individual objects can be specified at any level in the taxonomy and the relation between individuals and classes is one of set membership. Classes and individuals can have arbitrary properties and relations, which have generic or specific interpretations respectively. 

\begin{figure}
\includegraphics[width=0.8\textwidth]{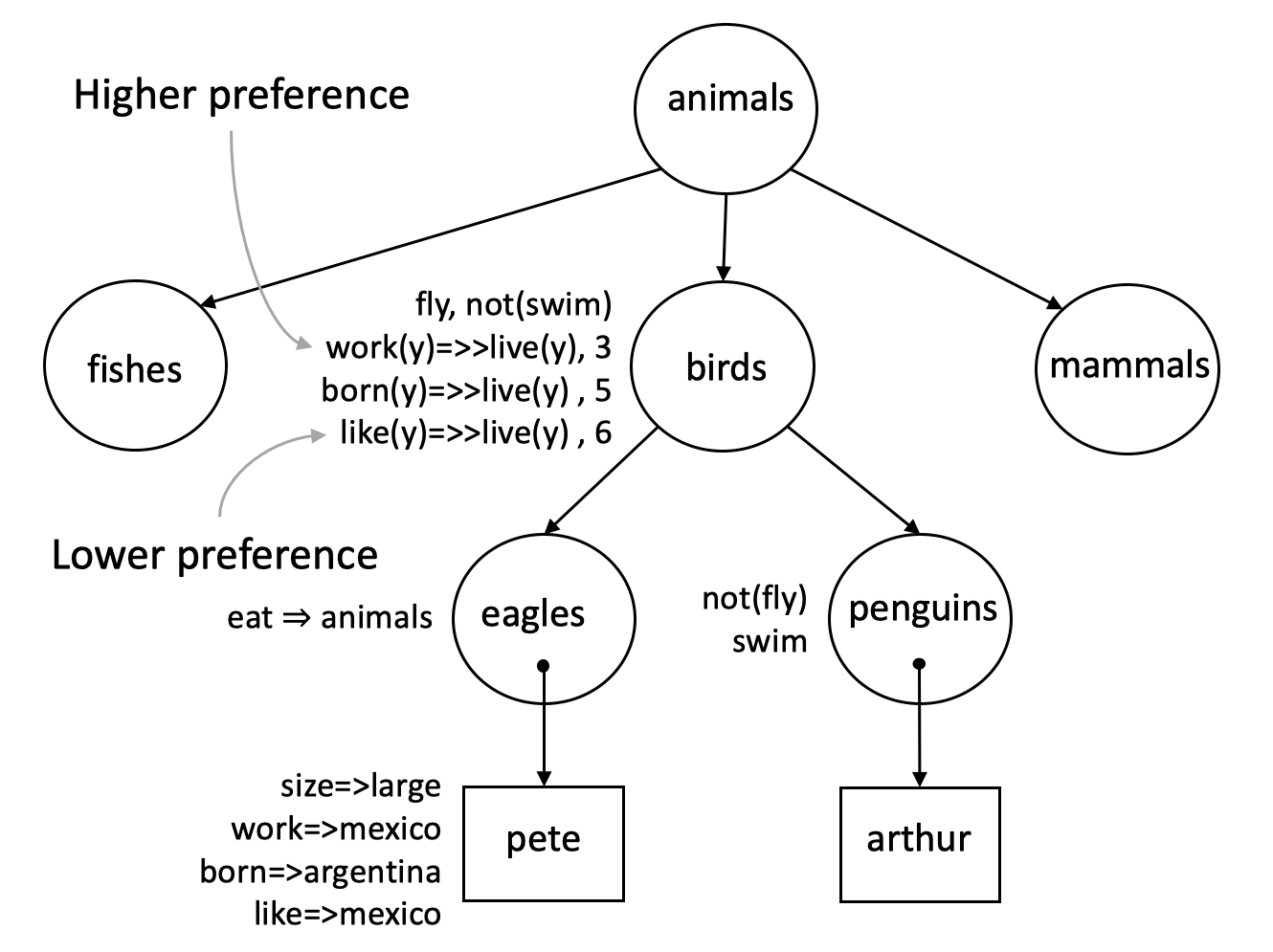}
\centering
\caption{Non-monotonic taxonomy.}
\label{fig:taxonomy}
\end{figure}

The taxonomy has a diagrammatic representation as illustrated in Figure \ref{fig:taxonomy}. Classes are represented by circles and individuals by boxes. The inclusion relation between classes is represented by a directed edge or arrow pointing to the circle representing the subordinated class, and the membership relation is represented by a bold dot pointing to the box representing the corresponding individual. Properties and relations are represented through labels associated to the corresponding circles and boxes; expressions of the form $\alpha$ $=>$ $\beta$ stand for a property or a relation -- where $\alpha$ stands for the name of the property or relation and $\beta$ for its corresponding value. The properties or relations are bounded within the scope of their associated circle or box. Classes and individuals can be also labeled with expressions of the form $\alpha$ $=>>$ $\beta$, $\gamma$ standing for implications, where $\alpha$ is an expression of the form $p_{0}, p_{1},...,p_{n}$ for $n \geq 0$, such that $p_{i} $ is a property or a relation, and $\beta$ stands for an atomic property or relation with a weight $\gamma$, such that $\beta$ holds for the corresponding class or individual with priority $\gamma$ if all $p_{i}$ in $\alpha$ hold. The KB service allows that objects of relations and values of properties are left underspecified augmenting its flexibility and expressive power.

For instance, the class \texttt{animals} at the top in Figure \ref{fig:taxonomy} is partitioned into \texttt{fishes}, \texttt{birds} and \texttt{mammals} where the class of birds is further partitioned into \texttt{eagles} and \texttt{penguins}. The label \texttt{fly} stands for a property that all birds have and can be interpreted as an absolute default holding for all individuals of such class and its subsumed classes. The label \texttt{eat $=>$ animals} denotes a relation between \texttt{eagles} with \texttt{animals} such that all eagles eat animals, and the question \emph{do eagles eat animals?} is answered \emph{yes}, without specifying which particular eagle eats and which particular animal is eaten. The properties and relations within the scope of a class or an individual --represented by circles and squares-- have such class or individual as the subject of the corresponding proposition, but these are not named explicitly. For instance, \texttt{like => mexico} within the box for Pete is interpreted as the proposition \emph{Pete likes Mexico}. In the case of classes such individual is unspecified but in the case of individuals it is determined. Likewise the labels \texttt{work(y) $=>>$ live(y),3}; \texttt{born(y) $=>>$ live(y),5}; and \texttt{like(y) $=>>$ live(y),6}; within the scope of \texttt{birds} stand for implications that hold for all unnamed individuals $x$ of the class \texttt{birds} and some individual \texttt{y}, which is the value of the corresponding property or the object of the relation --e.g. \emph{if x works at y then x lives at y}. Such implications are interpreted as conditional defaults, preferences or abductive rules holding for all birds that work at, were born in or like \texttt{y}. The integer numbers are the weights or priorities of such preference, with the convention that the lower the value the larger its priority. Labels without weights are assumed to have a weight of $0$ and represent the absolute properties or relations that classes or individuals have. The label \texttt{size $=>$ large} denotes the property \texttt{size} of \texttt{pete} and its corresponding value which is \texttt{large}. The labels \texttt{work $=>$ mexico}; \texttt{born $=>$ argentina}; and \texttt{like $=>$ mexico}; denote relations of Pete with their corresponding objects (M\'exico and Argentina). The system also supports the negation operator \texttt{not}, so all atoms can have a positive or a negative form (e.g., \texttt{fly}, \texttt{not(fly)}). 

Class inclusion and membership are interpreted in terms of the inheritance relation such that all classes inherit the properties, relations and preferences of their sub-summing or dominant classes, and individuals inherit the properties, relations and preferences of their class. Hence, the extension or closure of the KB is the knowledge specified explicitly plus the knowledge stated implicitly through the inheritance relation.

The addition of the \texttt{not} operator allows the expression of incomplete knowledge --as opposed to the Closed World Assumption (CWA). Hence queries are interpreted in relation to the strong negation and may be answered \emph{yes}, \emph{no} and \emph{not known}. For instance, the questions \emph{do birds fly?}, \emph{do birds swim?} and  \emph{do fish swim?} in relation to Figure \ref{fig:taxonomy} are answered \emph{yes}, \emph{no} and \emph{I don't know }.\footnote{In case the CWA were assumed queries would be right only in case complete knowledge about the domain were available, but could be wrong otherwise. For instance, the queries \emph{do fish swim?} and \emph{do mammals swim?} would be both answered \emph{no} in relation to the CWA, which would be wrong for the former but generally right for the latter.}

Properties, relations and preferences can be thought of as defaults that hold in the current class and over all the subsumed classes, as well as for the individual members of such classes. Such defaults can be positive, e.g. birds fly, but also negative, e.g. birds do not swim; defaults can have exceptions such as penguins that are birds that do not fly but swim. 

The introduction of negation augments the expressive power of the representational system, and allows for the definition of exceptions, but also allows the expression of contradictions, such as that penguins can and cannot fly, and swim and do not swim. To support this expressiveness and coherently reason about this kind of concepts, we adopt the principle of specificity which states that in case of conflicts of knowledge the more specific propositions are preferred. Subsumed classes are more specific than subsuming classes, and individuals are more specific than their classes. Hence, in the present example, the answer to \emph{do penguins fly?}, \emph{do penguins swim?} and \emph{does Arthur swim?} are \emph{no}, \emph{yes} and \emph{yes}.

The principle of specificity chooses a consistent extension of a set of atomic propositions, positive and negative, that can be produced out of the empty set by obtaining two extensions or branches, one with the positive and the other with its negation, one proposition at a time, for all end nodes of each branch and for all atomic propositions that can be formed with the atoms in the theory. These extensions give rise to a binary tree of extended theories in which each path represents a consistent theory but all different paths are inconsistent between each other. In the present example, the principle of specificity chooses the branch including \texttt{\{not(fly(pinguins)), swin(pinguins), not(fly(arthur)), swin(arthur)\}}. The set of possible theories that can be constructed in this way are referred to as \emph{multiple extensions} \cite{reiter-1980}.

The principle of specificity is a heuristics for choosing a particular consistent theory among all possible extensions. Its utility is that the extension at each particular state of the ontology is determined directly by the structure of the tree --or the strict hierarchy. Changing the ontology --augmenting or deleting classes or individuals, or changing their properties or relations-- changes the current theory; some propositions may change their truth value and some attributes may change their values, but the inference engine chooses always the corresponding consistent theory or the coherent extension.

Preferences can be thought of as conditional defaults that hold in the current class and over all subsumed classes, as well as for their individual members, if their antecedents hold. However, this additional expressiveness gives rise to contradictions or incoherent theories but this time due to the implication. In the present example the preferences  \texttt{work(y) $=>>$ live(y),3} and \texttt{born(y) $=>>$ live(y),5} of birds are inherited to Pete whom works in M\'exico but was born in Argentina; as Pete works in M\'exico and was born in Argentina, he therefore lives both in M\'exico and in Argentina, which is incoherent. This problem is solved by the present KB service through the weight value or priority, and as this is 3 for M\'exico and 5 for Argentina, the answer for \emph{where does Pete lives?} is \emph{M\'exico}.

Preferences can also be seen as abductive rules that provide the most likely explanation for an observation. For instance, if the property \texttt{live=>mexico} is added within the scope of Pete, the question of \emph{why does Pete live in M\'exico} can be answered \emph{because he works in M\'exico} --i.e., \texttt{work(y) $=>>$ live(y),3}--, which is preferred over the alternative  \emph{because he likes M\'exico} --i.e., \texttt{like(y) $=>>$ live(y),6}-- since the former preference has a lower priority. This kind of rules can also be used to diagnose the causes or reasons of arbitrary observations, and constitute a rich conceptual resource to deal with unexpected events that happen in the world.

The KB is specified as a list of Prolog clauses with five arguments: 1) the class id; 2) the subsuming or mother class; 3) the list of properties of the class; 4) the list of relations of the class; and 5) the list of individual objects of the class. Every individual is specified as a list, with its id, the list of its properties and the list of its relations. Each property and relation is also specified as a list including the property or relation itself and its corresponding weight. Thus, preferences of classes and individuals may be included in both the property list and the relation list, suggesting that they constitute conditional properties and relations. Id's, properties and relations are specified as attribute-value pairs, such that values can be objects of well-defined Prolog's forms. The actual code of the KB illustrated in Figure \ref{fig:taxonomy} is given in Listing \ref{list:code-taxonomy}.

\begin{listing*}[t]
\begin{verbatim}[frame=single,fontsize=\footnotesize]
[
 %The `top' class in mandatory
 class(top,none,[],[],[]), 
 class(animals,top,[],[],[]),
 class(fish,animals,[],[],[]),
 class(birds,animals,[[fly,0], 
 		     [not(swim),0], 
 		     [work=>'-'=>>live=>>'-',3],
		      [born=>'-'=>>live=>>'-',5],
		      [like=>'-'=>>live=>>'-',6]],
		[],[]),
 class(mammals,animals,[],[],[]),
 class(eagles,birds,[],[[eat=>animals,0]],
 		[[id=>pete,[[size=>large,0]],
				[[work=>mexico,0],
				 [born=>argentina,0],
				 [like=>mexico,0]
				]]
 		 ]),
 class(penguins,birds,[[swim,0],[not(fly),0]],[],[[id=>arthur,[],[]])
]
\end{verbatim}
  \vspace{-1em}
  \caption{Full code of example taxonomy.}
  \label{list:code-taxonomy}
\end{listing*}

 The KB service provides eight main services for retrieving information from the non-monotonic KB over the closure of the inheritance relations \cite{Non-Monotonic-KB-2017}, as follows: 
\begin{enumerate}
\item \texttt{class-extension(Class, Extension)} provides the set of individuals in the argument class. If this is \texttt{top}, this service provides the full set of individuals in the KB.
\item \texttt{property-extension(Property, Extension)} provides the set of individuals that have the argument property in the KB.
\item \texttt{relation-extension(Relation, Extension)} provides the set of individuals that stand as subjects in the argument relation in the KB.
\item \texttt{explanation\_extension(Property/Relation, Extension)} provides the set of individuals with an explanation supporting why such individuals have the argument property/relation in the KB.
\item \texttt{classes\_of\_individual(Argument, Extension)}: provides the set of mother classes of the argument individual.
\item \texttt{properties\_of\_individual(Argument, Extension)}: provides the set of properties that the argument individual has.
\item \texttt{relations\_of\_individual(Argument, Extension)}: provides the set of relations in which the argument individual stands as subject.
\item \texttt{explanation\_of\_individual(Argument, Extension)} provides the supporting explanations of the conditional properties and relations that hold for the argument individual.
\end{enumerate}

These services provide the full extension of the KB at a particular state. There are in addition services to update the values of the KB. There are also services to change, add or delete all objects in the KB, including classes and individuals, with their properties and relations. Hence the KB can be developed incrementally and also updated during the execution of a task, and the KB service provides always a coherent value. The full Prolog's code of the KB service is available at \url{https://bit.ly/non-monotonic-kb}. 

The KB services are manipulated by dialogue models as $SitLog$'s user functions. These services are included as standard $SitLog$'s programs that are used on demand during the interpretation of $SitLog$'s situations. Such services are commonly part of larger $SitLog$ programs representing input and output speech acts that are interpreted within structured dialogues defined through dialogue models. Hence, conceptual inferences made on demand during the performance of linguistic and interaction behavior constitute the core of our conceptual model of service robots.

\section{The Daily-Life Inference Cycle}
\label{sec:daily-life-inf-cycle}

Next we address the specification and interpretation of the daily-life inference cycle, as described in Section \ref{sec:introduction}. This cycle is studied from two different perspectives: the first consists of the pipe-line execution of a diagnosis, a decision-making and a planning inference, and involves the explicit definition of a problem space and heuristic search; the second is modelled through the interaction of appropriate speech-acts protocols and the extensive use of preferences. We refer to these two approaches as \emph{deliberative} and \emph{conceptual} inference strategies. The former is illustrated with a supermarket scenario, where the robot plays the role of an assistant, and the latter with a home scenario, where the robot plays the role of a butler, as described in Sections \ref{sec:deliberative-inference} and \ref{sec:conceptual-inference} respectively. The actors play analog roles in both settings, e.g., attending commands and information request related to a service task, and bringing objects involved in such requests or placing objects in their right locations, but each scenario emphasises a particular aspect of the kind of support that can be provided by service robots.

The robot behaves cooperatively and must satisfy a number of cognitive, conversational and task obligations, as follows:
\begin{itemize}
    \item Cognitive obligations ($CO$): 
    \begin{itemize}
        \item update its KB whenever it realizes that it has a false belief;
        \item notify the human user of such changes, so he or she can be aware of the beliefs of the robot;
    \end{itemize}
    \item Conversational obligations: to attend successfully the action directives or information requests expressed by the human user;
    \item Task obligations ($TO$): to position the misplaced objects in their corresponding shelves or tables.
\end{itemize}

The cognitive obligations manage the state of beliefs of the robot, and its communication to the human user. These are associated to perception and language, and are stated for the specific scenario. Conversational and task obligations may have some associated cognitive obligations too, that must be fulfilled in conjunction with the corresponding speech acts or actions. 

In both the supermarket and home scenarios there is a set of objects that belong to a specific class --e.g., food, drinks, bread, snacks, etc.-- and each shelf or table should hold objects of a designated class. Let $P_i = \{p_1,...,p_j\}$, $Q_i = \{q_1,...,q_j\}$ and $M_i = \{m_1,...,m_j\}$ be the sets of observed, unseen/missing and misplaced objects respectively on the shelf or the table $s_i$ in a particular observation at time $o_t = <s_i,P_t,Q_t,M_t>$ in relation to the current state of the KB. We assume that the behavior $see$ inspects the whole shelf or table in every single observation, and these three sets can be computed directly. $M_t \subseteq P_t$ must hold, and all objects in $P_t - M_t$ should belong to the class associated to the shelf $s_i$. 

Let $M_{{KB}_i}$ be the set of objects of the class $s_i$ that are believed to be misplaced in other shelves at the observation $o_t$ and $Misplaced_{KB}$ the full set of believed misplaced objects in the KB at any given time. Let $Missing_{o_t}$ be $Q_t - M_{{KB}_i}$; i.e., the set of objects of the shelf's class that the robot does not know where are placed at the time of the particular observation $o_t$.

Whenever an observation $o_t$ is made, the robot has the cognitive obligation of verifying whether it is consistent with the current state of the KB, and correct the false believes, if any, as follows:
\begin{enumerate}
    \item For every object in $Missing_{o_t}$ state the exception in the KB --i.e., that the object is not in its corresponding shelf; notify the exception, and that the robot does not know where such an object is!
    \item For every object in $M_t$ verify that the object is marked in the KB as misplaced at the current shelf; otherwise, update the KB accordingly and notify the exception.
\end{enumerate}

The conversational obligations are associated to the linguistic interaction with the human user. For instance, if he or she expresses a fetch command, the robot should move to the expected location of the requested object, grasp it, move back to the location where the user is expected to be, and hand the object over to him or her. The command can be simple, such as \emph{bring me a coke} or \emph{place the coke in the shelve of drinks}; or composite, such as \emph{bring me a coke and a bag of crisps}.

A core functionality of the GPSR is to interpret the speech acts in relation to the context and produce the appropriate sequence of behaviors, which is taken to be the meaning of the speech act. Such list of behaviors can also be seen as a schematic plan that needs to be executed to satisfy the command successfully. The general problem-solving strategy is defined along the lines of the GPSR as described above \cite{service-robot-2014}. 

The task obligations are generated along the execution of a task, when the world is not as expected and should eventually be fixed. For instance, the $see(object_i)$ behavior produces, in addition to its cognitive obligations, the task obligations of placing the objects in the sets $Q_t$, $M_t$ and $Missing_{o_t}$ in their right places. These are included in the list $Pending\_Task$.

All behaviors have a $status$ indicating whether the behavior was accomplished successfully or whether there was an error, and in this latter case, its type. Every behavior has also an associated manager that handles the possible termination status; if the status is $ok$ the behavior's manager finishes and passes the control back to the dialogue manager or the GPSR dispatcher.

However, when the behavior terminates with an error, the manager executes the action corresponding to the status type. There are two main cases: i) when the status can be handled with a recovery protocol and ii) when inference is required. An instance of case i) is the $move(shelf_j)$ behavior that may fail because there is a person blocking the robot's path, or a door is closed and needs to be opened. The recovery protocols may ask the person to move away and, in the latter situation, either ask someone around to open the door or execute the open-door behavior instead, if the robot does have such behavior in its behavior's library. An instance of case ii) is when the $take$ behavior --which includes a $see$ behavior-- fails to find the object in its expected shelf. This failure prompts the execution of the daily-life inference cycle.

\section{Deliberative Inference}
\label{sec:deliberative-inference} 

This inference strategy is illustrated with the supermarket scenario in Figure \ref{fig:supermarket-scenario}. This has the following elements:

\begin{enumerate}
    \item The supermarket consists of a finite set of shelves $S = \{s_1,...s_n\}$ at their corresponding locations $L = \{l_1,...,l_n\}$, each having an arbitrary number of objects or entities $O = \{o_1,...,o_n$\} of a particular designated class $c_i \in C$, the set of classes; for instance, $C = \{drinks, food, bread\}$;
    \item The human client, who may require assistance;
    \item The robot, that has a number of conversational, task and cognitive obligations;
    \item A human supermarket assistant who's job is to bring the products from the supermarket's storage and place them on their corresponding shelves.
\end{enumerate}

The cognitive, conversational and task obligations are as stated above. A typical command is \emph{bring me a coke}, which is interpreted as [$acknowledge$,  $grasp(coke)$, $find(user)$, $deliver(coke,user)$], where $grasp(coke)$ is the composite behavior $kb\_get\_shelf\_of\_object(object_i,shelf_j)$, $move(shelf_j)$, $find(object_i)$ and $take(object_i)$.

In this scenario the priority is to satisfy the customer, and a sensible strategy is to achieve the $take(object_i)$ action as soon as possible and complete the execution of the command, and use the idle time to carry on with the $Pending\_Task$. These two goals interact and the robot may place some misplaced objects along the way if the actions deviate little from the main conversational obligation. If the sought object is placed at its right shelf the command can be performed directly; otherwise, the robot must engaged in the daily-life inference cycle to find the object, take it and deliver it to the customer. These conditions are handled by the behavior's manager of the behavior $take$ --which in turn uses he behavior $see$, with its associated cognitive obligations.

The arguments of the inference procedure are:

\begin{enumerate}
    \item The current $take(object_i)$ behavior;
    \item The list $Previous\_Shelves$ of shelves already inspected by the robot including the objects placed on them --that corresponds to the states of the shelves as arranged by the human assistant when the scenario was created, as discussed below in \ref{subsec:diagnosis}; this list is initially empty;
    \item The set $Objects\_Placed$ of objects already put in their right locations by previous successful place actions performed by the robot in the current inference cycle; this set is initially empty.
\end{enumerate}

The inference cycle proceeds as follows:
\begin{enumerate}
    \item Perform a diagnosis inference in relation to the actual observations already made by the robot; this inference renders the assumed actions made by the human assistant when he or she filled up the stands including the misplaced objects $Misplaced_{KB}$;
    \item Compute the $Decision$ in relation to the current goal --e.g., $take(object_i)$--  and possibly other subordinated place actions in the current $Pending\_Task$;
    \item Induce the plan consisting of the list of behaviors $Plan$ to achieved $Decision$;
    \item Execute the plan in $Plan$; this involves the following actions:
    \begin{enumerate}
        \item update $Previous\_Shelves$ every time the robot sees a new shelf;
        \item update the KB whenever an object is placed on its right shelf, and accordingly update the current $Pending\_Task\_Obligations$; and update $Objects\_Placed$;
        \item if the $object_i$ is not found at its expected shelf when the goal $take(object_i)$ is executed, invoke the inference cycle recursively with the same goal and the current values of $Previous\_Shelves$ and $Objects\_Placed$ which may not be empty.
    \end{enumerate}
\end{enumerate}

\subsection{Diagnosis Inference}
\label{subsec:diagnosis}

The diagnosis inference model is based on a number of assumptions that are specific to the task and the scenario, as follows:

\begin{enumerate}
    \item The objects --e.g., drinks, food and bread products-- were placed in their corresponding shelves by the human assistant who can perform the actions $move(s_{i})$ --move to $location$ $l_{i}$ of $shelf$ $s_{i}$ from its current location-- and $place(o_{i})$ --i.e., place $object_{i}$ on the shelf at the current location. The assistant can start the delivery path at any of arbitrary shelf, can carry as many objects as needed in every move action, and he or she places all the objects in a single round.
    \item The believed content of the shelves is stored in the robot's KB. This information can be provided in advanced or by the human assistant through a natural language conversation, which may be defined as a part of the task structure.
    \item If an object is not found by the robot in its expected shelf, it is assumed that it was originally misplaced in another shelf by the human assistant. Although in actual supermarkets there is an open-ended number of reasons for objects to be misplaced, in the present scenario this is the only reason considered.
    \item The robot performs local observations and can only see one shelf at a time, but it sees all the objects on the shelf in a single observation.
\end{enumerate}

The diagnosis consists of the set of moves and placing actions that the human assistant is assumed to have performed to fill up all the unseen shelves given the current and possibly previously observed shelves. Whenever there are mismatches between the state of the KB and the observed world, a diagnosis is rendered by the inferential machinery.\footnote{It should be considered that even the states of observed shelves are also hypothetical as there may have been visual precision and/or recall errors --i.e, objects may have been wrongly recognized or missed out; however, when this happens the robot can recover only later on when it realizes that the state of the world is not consistent with its expectations, and has to reconsider previous diagnoses.}

The diagnosis inference is invoked when the $see(object_j)$ at shelf $s_i$ within the $take(object_j)$ behavior fails. The KB is updated according to the current observation, and contains the beliefs of the robot about the content of the current and the remaining shelves. The current observation $see(object_j)$ renders the set of missing objects $Missing_{o_t}$ and $M_t$ of missing and misplaced objects at the current shelf $s_i$. If the sought object is of the class $c_i$ of the shelf, it must be within $Missing_{o_t}$ or the supermarket has run out of such object; otherwise the robot believed that the sought object was already misplaced in a shelf of a different class $c_j$, but the failed observation showed that such belief was false. Consequently the sought object must be included in $Missing_{o_t}$ and the KB must be updated with the double exception in the KB: that the object is not in the current shelf --and was not in its corresponding shelf-- and hence must be in one of the shelves that remain to be inspected in the current inference cycle. This illustrate that negative propositions increases the knowledge productively, as the uncertainty is reduced.

The diagnosis procedure involves extending the believed content of all unseen shelves $S_k$ with the content of $Missing_{o_t}$, avoiding repetitions. The content of the shelves seen in previous observations is already known.

There are many possible heuristics to make such an assignment; here we simply assume that $s_j$ is the closest unseen shelf --in metrical distance-- to the current shelf $s_i$, and distribute the remaining objects of $Missing_{o_t}$ in the remaining unseen shelves randomly. The procedure renders the known state of shelf $s_i$ --unless there were visual perception errors-- and the assumed or hypothetical states of the remaining unseen shelves. 

The diagnosis is then rendered directly by assuming that the human assistant moved to each shelf and placed on it all the objects in its assumed and known states. There may be more than one known state because the assumption made at a particular inference cycle may have turned out wrong, and the diagnosis may have been invoked with a list of previous observed shelves whose states are already known. 

\subsection{Decision-Making Inference}
\label{subsec:decision}

In the present model deciding what to do next depends on the task obligation that invoked the inference cycle in the first place $TO$, e.g., $take(object_i)$, and the current $Pending\_Task$. Let the set $Potential\_Decisions = TO \cup Pending\_Task$. Compute the set $Potential\_Decisions_{subsets}$ consisting of all subsets of $Potential\_Decisions$ that include $TO$.

The model could also consider other parameters such as the mood of the customer or whether he or she is in a hurry, that can be thought of as constraints in the decision-making process; here we state a global parameter $r_{max}$ that is interpreted as the maximum cost that can be afforded for the completion of the task.

We also consider that each action performed by the robot has an associated cost in time --e.g., the parameters associated to the behaviors $take$ and $deliver$-- and a probability to be achieved successfully --e.g., the parameters associated to a $move$ action. The total cost of an action is computed by a restriction function $r$.

The decision-making module in relation to $Potential\_Decisions_{subsets}$ proceeds as follows:
\begin{enumerate} 
    \item Compute the cost $r_i$ for all sets in $Potential\_Decisions_{subsets}$; 
    \item $Decisions$ is the set with maximal cost $r_i$ such that $r_i \leq r_{max}$.
\end{enumerate}

\subsection{Plan Inference}
\label{subsec:planning}

The planning module searches the most efficient way to solve a set of $CO$ and $TO$. Each element of $CO \cup TO$ implies a realignment in the position of the objects in the scenario, either carrying and object to another shelf or delivering it to a client.   

Each $TO$ is transformed in a list of basic actions of the form:
\\
$[move(s_a), take(o_k), move(s_b), deliver(o_k)]$
\\
and each $CO$ is transformed in a list of basic actions of the form:
\\
$[move(s_a), take(o_k), search(client), deliver(o_k)]$
\\
where $s_a$ is the shelf containing the object $o_k$ according to the diagnosis module, and $s_b$ is the correct shelf where $o_k$ should be according to the KB. All the lists are joined in a multiset of basic actions $B$.    

The initial state $S_0$ of the search tree contains:
\begin{itemize}
    \item The current location of the robot ($l_k$)
    \item The actual state of the right hand (free or carrying the object $o_r$).
    \item The actual state of the left hand (free or carrying the object $o_l$).
    \item The list $R$ of remaining $CO \cup TO$ to solve.
    \item The multiset $B$ of basic actions to solve the elements in $R$.
    \item The list of basic actions of the plan $P$ (in this moment is still empty). 
\end{itemize}

The initial state is put on a list $F$ of all the not expanded nodes in the frontier of the tree. The search algorithm proceeds as follows: 
\begin{enumerate}
    \item Select one node to expand from $F$. The selection criteria of the node of $F$ is DFS. The cost and probability of each action in the current plan P in the node is used to compute a score.  
    \item When a node $S_i$ has been selected, a rigorous analysis of $B$ is performed. For each basic action in B, check if the following preconditions are satisfied:
    \begin{itemize}
        \item Two subsequent navigation moves are banned. If the action is a $move$ or a $search(user)$, discard if the last action of $P$ is a $move$ or a $search(user)$.
        \item Only useful observations. If the action is a $search(object)$, discard if the last action of $P$ is a $search(object)$, $search(user)$, or if the robot actually has objects in both hands. 
        \item Only deliveries after taking. If the action is $deliver(o_i)$, the action $take(o_i)$ should be included previously in the plan.
         \item Only take actions if at least one hand is free. 
    \end{itemize}
    \item For each basic actions of $B$ not discarded using the preconditions generate a successor node $S_{ij}$ in this way: 
    \begin{itemize}
        \item If the basic action is $move(s)$ or $search(user)$ change the current location of the robot to $s$ or the user position respectively. If not, the current location of the robot in $S_{ij}$ is the same as $S_i$.
        \item Update the state of the right and left hand if the basic action is a $take$ or a $deliver$.
        \item If the basic action was a $deliver$, delete the associated element in the list $R$ of remaining $CO \cup TO$. If the list gets empty, a solution has been found. 
        \item Remove the basic action used to create this node from $B$.
        \item Add the basic action to the plan $P$.
    \end{itemize}
    \item Return to step 1 to select a new node.
\end{enumerate}

When a solution has been found in the tree the plan $P$ is post processed to generate a list of actions specified in terms of $SitLog$ basic behaviors, which can be used by the dispatcher.

A video showing a demo of the robot Golem-III performing as a supermarket assistant, including all the features described in this section is available at \url{http://golem.iimas.unam.mx/inference-in-service-robots}. The KB-system and the full Sitlog's code are also available at \url{https://bit.ly/deliberative-inference}.

\section{Conceptual Inference}
\label{sec:conceptual-inference}

This inference strategy is illustrated with a home scenario in which the robot plays the role of a butler, as follows:

\begin{enumerate}
    \item The home has a number of rooms $R = \{r_1,...r_n\}$; a finite set of shelves $S = \{s_1,...s_n\}$ at their corresponding locations $L = \{l_1,...,l_n\}$ --that may be within any arbitrary room, each having an arbitrary number of objects $\{o_1,...,o_n\}$ of a particular designated class $c_i \in C$, the set of classes;
    \item The human user, who may require assistance;
    \item The human user has a child that plays at home and misplaces objects in wrong tables;
    \item The robot has a number of conversational, task and cognitive obligations, that are the same as in the supermarket scenario, but in addition it may provide the user plausible explanations about the causes of unexpected observations;
    \item The user has a number of preferences that are stored in the robot's KB, which can be changed dynamically; in addition, the robot may know about general health recommendations that can be deployed along the execution of a task.
\end{enumerate}

The details of the implementation of the home scenario, including the use of the conditional defaults in the KB and the way they are managed, are explained in Appendix \emph{Conceptual Inference Scenario}. The roles played by the agents and the speech acts involved are analogous to those of the supermarket scenario. Additionally, in the home scenario the robot is familiar with the human user, has knowledge about his or her preferences and/or habits, and may use this knowledge to assist him or her better. The robot's behavior is proactive and the task relies on a more extensive use of cooperative dialogues, where the robot may have conversational initiatives. A task oriented conversation in this setting between the robot Golem-III and a human user named Luis is as follows:
\begin{enumerate}
    \item \textbf{Golem}: \textit{Hi Luis.}
    \item \textbf{Golem}: \textit{Please tell me what do you like best: malz or coke?}
    \item \textbf{Luis}: \textit{Hi Golem, I like malz.}
    \item \textbf{Golem}: \textit{Excellent! Malz is much healthier than the coke!}
    \item \textbf{Golem}: \textit{What do you like best: noodles or biscuits?}
    \item \textbf{Luis}: \textit{I like noodles best.}
    \item \textbf{Golem}: \textit{Great! I will recall your choice!}
    \item \textbf{Golem}: \textit{Do you have any more preferences?}
    \item \textbf{Luis}: \textit{No thanks Golem; that is ok.}
    \item \textbf{Golem}: \textit{Thank you Luis.}
 \end{enumerate}
 At this point Luis goes to work and takes some time to be back; when he does Golem meets him at the door.
 \begin{enumerate}
    \setcounter{enumi}{10}
    \item \textbf{Golem}: \textit{Hi Luis, how was your day?}
    \item \textbf{Luis}: \textit{I had a bad day!}
    \item \textbf{Golem}: \textit{Sorry to hear that; do you want me to do something for you?}
    \item \textbf{Luis}: \textit{Bring me something to drink and biscuits.}
    \item \textbf{Golem}: \textit{Ok. I will bring you malz, your favorite drink.}
    \item \textbf{Luis}: \textit{Yes.}
    \item \textbf{Golem}: \textit{But you like noodles better than biscuits! Shall I bring you the biscuits?}
    \item \textbf{Luis}: \textit{No Golem, noodles is ok.}
    \item \textbf{Golem}: \textit{Ok. I will bring you the malz and the noodles. I will get the malz.}
\end{enumerate}   
Golem goes to the expected locations of the offered objects since they should be at their respective shelves, although there may be exceptions. So Golem reaches the shelf of drinks where the malz should be placed and finds it.
\begin{enumerate}
\setcounter{enumi}{19}
    \item \textbf{Golem}: \textit{Attempting to grab the malz with my left arm.}
\end{enumerate}
Golem grabs the malz.
\begin{enumerate}
\setcounter{enumi}{20}   
    \item \textbf{Golem}: \textit{I took the malz.}
    \item \textbf{Golem}: \textit{I will get the noodles.}
\end{enumerate}  
Golem goes to the expected location of noodles, which is the table for food; however, the noodles are not there.
\begin{enumerate}
    \setcounter{enumi}{22}
    \item \textbf{Golem}: \textit{The noodles is not in the shelf of food.}
\end{enumerate}
Golem goes to the next preferred shelf for noodles, which is snacks; the noodles are found there.
\begin{enumerate}
    \setcounter{enumi}{23}
    \item \textbf{Golem}: \textit{Attempting to grab the noodles with my right arm.}
\end{enumerate}   
Golem grabs the noodles;
\begin{enumerate}
    \setcounter{enumi}{24}
    \item \textbf{Golem}: \textit{I took the noodles.}
\end{enumerate}
Golem goes to the location where Luis is expected to be, which is the living room since he is back from work and tired after a bad day. Golem finds him there.
\begin{enumerate}
    \setcounter{enumi}{25}  
    \item \textbf{Golem}: \textit{Here is the noodles.}
\end{enumerate}  
    Golem handles the noodles to Luis.
\begin{enumerate}
    \setcounter{enumi}{26}   
    \item \textbf{Golem}: \textit{Here is the malz.}
\end{enumerate}
Golem handles the malz to Luis.
\begin{enumerate}
    \setcounter{enumi}{27} 
    \item \textbf{Golem}: \textit{I found the noodles in the shelf of snacks but it should be in the shelf of food; do you want me to change the preferred location of noodles to the shelf of snacks?}
    \item \textbf{Luis}: \textit{Yes Golem, please.}
    \item \textbf{Golem}: \textit{Ok. I updated my KB with your new preference.}
    \item \textbf{Golem}: \textit{I also noticed that the coke is not in its right place.}
    \item \textbf{Golem}: \textit{I think that the explanation for this is that the coke was misplaced there by your child.}
    \item \textbf{Golem}: \textit{Do you want me to take it to its right shelf?}
    \item \textbf{Luis}: \textit{Yes Golem, please.}
    \item \textbf{Golem}: \textit{Ok. I will take it to the shelf of drinks.}
 \end{enumerate}
 Golem goes to the shelf of snacks and takes the coke.
 \begin{enumerate}
    \setcounter{enumi}{35}
    \item \textbf{Golem}: \textit{Attempting to grab the  coke  with my right arm.}
    \item \textbf{Golem}: \textit{I took the  coke.}
\end{enumerate}
Golem goes to the shelf of drinks, places the coke there and goes back to the living room with Luis.
 \begin{enumerate}
    \setcounter{enumi}{37}
    \item \textbf{Golem}: \textit{I put the coke in its right shelf.}
    \item \textbf{Golem}: \textit{All the objects are placed in their right shelves.}
    \item \textbf{Golem}: \textit{The task is finished.}
    \item \textbf{Golem}: \textit{Good bye.}
\end{enumerate}
This scenario could be implemented with the $GPSR$ too; however, here we use an alternative strategy that offers an additional perspective of the framework. This is based on the direct specification of speech act protocols defined in $SitLog$. These are intentional structures in which performing a speech act establishes a number of conversational obligations that must be fulfilled before the dialogue proceeds to the next transaction. For instance, a command must be executed, and a question must be answered. The dialogue models are designed considering the user's preferences and the whole task oriented conversation is modelled as the interpretation of one main protocol that embeds the goals of the task. The design of the dialogue models is loosely based on the notion of balanced transactions of the DIME-DAMSL annotation scheme \cite{dime-damls}. 

In the first section of the dialogue from (1) to (10) the robot asks for the user's preferences, and the KB is updated accordingly. The interpretation considers the user' utterances in relation to his or her current preferences, and also in relation to other generic preferences that are stated in advance in the KB.

Utterances (11) to (19) consist on a speech act protocol to make and accept an offer. The protocol starts with greeting an open offer expressed by the robot in (11-13), that is answered with a user's request in (14); however, this is under-specified and vague; the robot resolves it using the user's preferences --his favorite drink-- but also by contrasting the ordered food with the user's own food preferences, which results in a confirmation question in (17). The user changes his request and the robot confirms the whole command in (18-19).

The robot executes the command with the corresponding embedded actions from (20) to (27). At this point a new protocol is performed from (28) to (30) due to the task obligation that was generated when the robot noticed that an object --the noodles-- was not placed on its corresponding shelf, and asks for the confirmation of a user's preference.

Then, another protocol to deal with a new task obligation is performed from (31) to (32), including the corresponding embedded actions. This protocol involves an abductive explanation, that is performed directly on the basis of the observation and the preference rule used backwards, as explained above in Section \ref{sec:kb-system}. This protocol is concluded with a new offer that is accepted and confirmed in (33-35). The new task is carried out as reported in (36-38). The task is concluded with the final protocol performed from (39) to (41).

\begin{figure}[tb]
\includegraphics[width=1.0\textwidth]{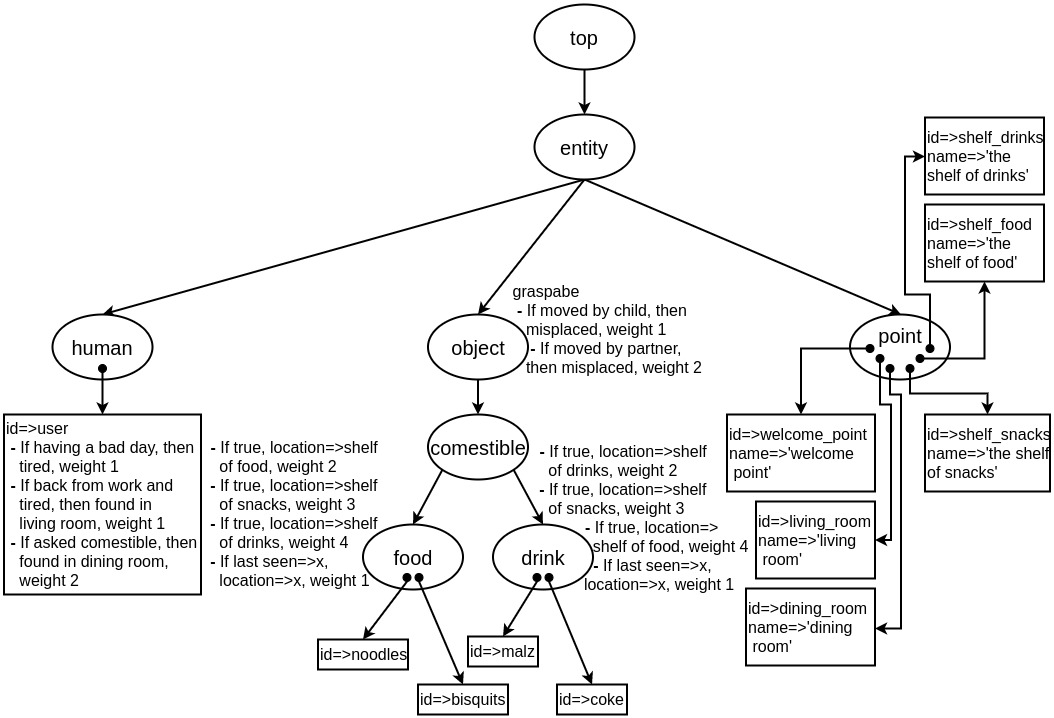}
\centering
\caption{KB with preferences.}
\label{diag:kb}
\end{figure}

The speech acts and actions performed by the robot rely on the state and dynamic evolution of the knowledge. The initial KB supporting the current dialogue is illustrated in Figure~\ref{diag:kb} and its actual code available at Listing~\ref{list:KB_conceptual_inf}. In it, the preferences are written as conditional defaults (e.g., \texttt{bad\_day=>>tired}), which are considered to successfully interact with the user and to achieve abductive reasoning. As the demo is performed some new elements are defined in the KB, such as the properties \texttt{back\_from\_work} and \texttt{ask\_comestible} added to the individual \texttt{user}. Later in the execution of the task, such properties play an important role determining the preferences of the user. 
%%%
% Start figure
%%%
%%%
% End figure
%%%

%%%
% Start listing
%%%
\begin{listing*}[tb]
\begin{verbatim}[frame=single,fontsize=\footnotesize]
[ class(top,none,[],[],[]), class(entity, top, [], [], []), 
class(human, entity, [], [],     [[id=>user, [ [bad_day=>>tired,1],
     [[back_from_work,tired]=>>found_in=>living_room,1], 
     [asked_comestible=>>found_in=>dining_room,2] ], []]]),
class(object, entity, [ [graspable=>yes,0],
  [moved_by=>child=>>misplaced,1], [moved_by=>partner=>>misplaced,2] ], [], []),
class(comestible, object, [], [], []),
class(food, comestible, [ ['-'=>>loc=>shelf_food,2],
     ['-'=>>loc=>shelf_snacks,3],['-'=>>loc=>shelf_drinks,4],
     [last_seen=>'-'=>>loc=>'-',1] ], [],
         [[id=>noodles, [], []], [id=>bisquits, [], []]]),
class(drink, comestible,[ ['-'=>>loc=>shelf_drinks,2],
     ['-'=>>loc=>shelf_snacks,3],['-'=>>loc=>shelf_food,4],
     [last_seen=>'-'=>>loc=>'-',1] ], [], 
         [[id=>coke, [], []], [id=>malz, [], []]]),
class(point, entity, [], [],[
     [id=>welcome_point,[[name=>'welcome_point',0]],[]],
     [id=>living_room, [[name=>'living room',0]],[]],
     [id=>dining_room, [[name=>'dining room',0]],[]],
     [id=>shelf_food,  [[name=>'the shelf of food',0]],[]],
     [id=>shelf_drinks,[[name=>'the shelf of drinks',0]],[]],
     [id=>shelf_snacks,[[name=>'the shelf of snacks',0]],[]]
]) ]
\end{verbatim}
\vspace{-1em}
\caption{KB with preferences.}
\label{list:KB_conceptual_inf}
\end{listing*}
%%%
% End listing
%%%

The daily-life inference cycle is also carried on in this scenario, although it surfaces differently from its explicit manifestation as a pipe-line inference sequence. 

As in the deliberative scenario, a diagnosis inference emerges when the expectations of the robot are not met in the world, although in the present case such a failure creates a task obligation that will be fulfilled later, such as in (28) and (31-33). However, instead of producing the whole set of actions that lead to the observed state, the robot focuses only on the next action or on producing the abductive explanation directly from the observed fact and the corresponding KB-Service, as in (32). 

In this setting there is also an implicit diagnosis that is produced from continuously verifying  whether there is a discrepancy between the user's manifested beliefs and intentions, and the preferences in the KB. For instance, this form of implicit diagnosis underlies utterances (5) and (17).

The decision making in this setting is also linked to the conversational structure and the preferences in the KB. Decisions are made on the basis of diagnoses, and have the purpose of reestablishing the desired state of the world, or to make the world and the KB consistent with the known preferences, and the robot makes suggestions to the user who is the one who makes the actual decisions as in (28-29) and (33-34).

The planning inference is also implicit, as the robot has the obligation to perform the action that conforms with the preferences, as when it inspects the shelves looking for objects in terms of their preferred locations, as in (23) and its associated previous and following actions.

The conceptual inference strategy relies on an interplay between the structure of speech acts transactions and the preferences stored in the KB, and avoids the explicit definition of a problem space and heuristic search. The inferences are sub-optimal, and rely on the conversational structure, a continuous interaction between the language and the KB, and the interaction with the world.

A video showing a demo of the robot Golem-III as a home assistant performing the task oriented conversation (1-41) is available at \url{http://golem.iimas.unam.mx/inference-in-service-robots}. The corresponding KB and $SitLog$ dialogue models are available at \url{https://bit.ly/conceptual-inference}.

\section{Conclusions and further Work}
\label{sec:conclusions}

In the paper we have reviewed the framework for specification and development of service robots that we have developed over the last years. This framework includes a conceptual model for service robots, a cognitive architecture to support it, and the $SitLog$ programming language for the declarative specification and interpretation of robotics task structure and behaviors. This language supports the definition of speech acts protocols that the robot performs during the execution of the task, fulfilling implicitly the objectives of goal-oriented conversations.

We have also presented a non-monotonic knowledge-base system for the specification of terminological and factual knowledge in robotics applications. The approach consists on the definition of an strict taxonomy that support defaults and exceptions, that can be updated dynamically. Conflicts of knowledge are resolved through the principle of specificity, and contingent propositions have an associated weight. The system allows the specification of preferences that are employed in the reasoning process and can provide plausible explanations about unexpected facts that are realized by the robot while performing the task.

The present framework allows us to model service robotics tasks through the definition of speech acts protocols; these protocols proceed while the expectations of the robot are met in the world. However, whenever no expectation is satisfied in a particular situation, the ground is lost, the robot gets out of context, and cannot proceed with the task. Such a contingency is met with two strategies: the first consists on invoking a recovering protocol, whose purpose is to restore the ground through interacting with other agents or the world; the second consists on resorting to symbolic reasoning --or thinking-- by invoking and executing the daily-life inference cycle. 

This cycle is studied through two different approaches: the first consists on the pipe-line implementation of a diagnosis, a decision making and a planning inference, and involves the explicit definition of a problem-space and heuristics search; and the second consists on the definition of the tasks in terms of speech act protocols that are carried on cooperatively between the robot and the human user, in which the ground is kept through the intense use of preferences stored in the robot's KB which are deployed along the robotics tasks. These approaches are called here deliberative and conceptual inference respectively.

We illustrated these two approaches with two fully detailed scenarios and showed how these are deployed in real-time in a full autonomous manner by the robot Golem-III. In the former the robot performs as a supermarket assistant and in the latter as a butler at home.

The deliberative inference scenario is structured along the lines of the traditional symbolic problem-solving strategy, and renders explicitly the three main stages of the daily-life inference cycle. Inference is conceived as a problem of optimization, where the chosen diagnosis, decisions and plans are the best solutions that can be found given the constraints of the task. The methodology is clear and highlights the different aspects of inference.

However, the three kinds of inferences are carefully designed and programmed beforehand; the methods and algorithms are specific to the domain; and it is unlikely that a general and domain independent set of algorithms can be developed. The method adopts a game playing strategy, and the interaction with the human-user is reduced to listen to the commands, and performing them in long working cycles. The conversational initiative is mostly on the human side, and the robot plays a subordinated role. For these reasons, although the strategy may yield acceptable solutions, it is somehow unnatural, and reflects poorly the strategy employed by people facing this kind of problems in similar kind of environments.

The conceptual strategy carries on with the three kind of inference but implicitly, based on informed guesses that use the preferences stored in the KB. In this latter approach the ground is not broken when the robot realizes that the world in not as expected, and the robot does not perform an explicit diagnosis, decision making and planning inferences; the focus is rather on what is the closer world or situation in which the current problem can be solved and act accordingly.

This approach renders much richer dialogues than the pipe-line strategy, where the inference load is shared between the two agents. The robot makes informed offers on the basis of the preferences, and the human user makes the choices; but also the robot can make choices, that may be confirmed by the user, and the robot takes conversational initiatives to a certain extent. Overall, the task is deployed along a cooperative conversation through the deployment of a speech acts protocol, that makes an intensive use of the knowledge and preferences stored in the KB, and the goals of the task are fulfilled as a co-lateral effect of carrying on with such protocols. In this approach the robot does not defines a dynamic problem space, and limits greatly heuristic search, as the uncertainty is captured in the preference.

Although at the present time the speech act protocols are specific for the task, we envisage the development of generic protocols that can be induced from the analysis of task oriented conversation, that can be instantiated dynamically with the content stored in the KB and the interaction with the world, and the approach can be made domain independent to a larger extent than the present one. For instance, by providing abstract speech acts protocols for making offers, information or action requests, etc. However, for the moment this enterprise is left for further work.

\section*{Acknowledgements}
The authors thank Iv\'an Torres, Dennis Mendoza, Caleb Rasc\'on, Ivette V\'elez, Lisset Salinas and Ivan Meza for the design and implementation of diverse robotic algorithms and to Mauricio Reyes and Hernando Ortega for the design and construction of the robot's torso, arms, hands, neck, head and face, and the adaptation of the platform. We also thank Varinia Estrada, Esther Venegas and all the members of the Golem Group who participated in the demos of the Golem robots over the years, and also to those who have attended the RoboCup competitions since 2011.

\appendix
\section{An example program in SitLog}
\label{appendix-1-sitlog}

In order to show the expressive power of SitLog the full code of the program in Figure \ref{fig:dummy} is provided. A DM is defined as a clause with three arguments as follows. 

\begin{Verbatim}[fontfamily=courier]
diag_mod(id(Arg_List), Situations, Local_Vars).
\end{Verbatim}

The first argument is an atom (i.e., the DM's name or Id) or a predicate in which case the functor is the DM's id and the list arguments are the arguments of the DM, which are visible within its body; the second is the list of situations; and the third the list of local variables. A situations is defined as a set of attribute-value pairs as was mentioned; situation id's need not be unique, and different instances of the same situation with the same id but different arguments or values can be defined.

Listings~\ref{list:dummy_main} and \ref{list:dummy_wait} include the clauses with the definitions of  \emph{main} and \emph{wait} of the program illustrated in Figure \ref{fig:dummy}.
The value of the input pipe is initialized by the value provided in the first occurrence of \texttt{out\_arg} and the global variables are declared as a global parameter of the application as follows:

\begin{Verbatim}[fontfamily=courier]
Global_Vars = [g_count_fs1 ==> 0,
               g_count_fs2 ==> 0].
\end{Verbatim}

%%%%%%%%%%%%%%%%%%%%%%
%% Listing 3
%%%%%%%%%%%%%%%%%%%%%%
\begin{listing}[tb]
\begin{Verbatim}[fontfamily=courier,fontsize=\footnotesize,frame=single,framesep=2mm]
% Main Dialogue Model
diag_mod(main,
 %Second argument: List of Situations
 [% Initial situation
  [id ==> is,
   type ==> speech,
   in_arg ==> In_Arg,
   out_arg ==> apply(when(If,True,False),
               [In_Arg=='monday','tuesday','monday']),
   % Local program
   prog ==> [inc(count_init,Count_Init)],
   arcs ==>
        [% Examples of Grounded forms
         finish:screen('Good Bye') => fs,
         % Example of predicate expectation and action
         [day(X)]:[date(get(day,Y)),  next_date(set(day,X))] => is,
         % Example of functional specification of
         % expectation, action and next situation
         [get(day,Day),apply(f(X),[In_Arg])]: [apply(g(X),[_])] =>
                       apply(h(X,Y),[In_Arg,Day])
        ]
  ],
  % Second Situation
  [id ==> rs,
   type ==> recursive,
   prog ==> [inc(count_rec, Count_Rec)],
   embedded_dm ==> sample_wait,
   arcs ==> [fs1:screen('Back to initial sit') => is,
             fs2:screen('Cont. recursive sit') => rs]
  ],
  % Final Situation
  [id ==> fs, type ==> final]
 ], % End list of situations
 % Third Argument: List of Local Variables
 [day ==> monday, count_init ==> 0, count_rec ==> 0]
). %End DM (main)
\end{Verbatim}
\caption{SitLog's Specification of the DM {\em main}.}
\vspace{-5pt}
\label{list:dummy_main}
\end{listing}
%%% End listing

The DM \emph{main} includes the list of the three situations (see Figure \ref{fig:dummy})\footnote{In this SitLog program neither the DM nor the situations use parameters; these are illustrated in the DMs representing the demos task structure and behaviors.}. The situation \texttt{is} has an application specific type, in this case \texttt{speech}. There is a specific perceptual interpreter for all user-defined types. These interpreters specify the interface between the expectations of the situation and the low level recognition processes. The type \texttt{speech} specifies that the expectation of the situation will be input through the speech modality. The notation of expectations is defined in the corresponding perceptual interpreter which instantiates the current expectations and returns the one satisfied in the current interpretation situation.

The \texttt{is} situation includes a local program -- defined by the \texttt{prog} attribute -- consisting of the SitLog's operator \texttt{inc} that increases by one the value of the local variable \texttt{count\_init} each time the situation is visited during the execution of the \emph{main} DM. Its \texttt{arcs} attribute is a list with the specification of its three exit edges. Each exemplifies a kind of expectation: a concrete one (i.e., \texttt{finish}), a list including one open predicate (i.e., \texttt{day(X)]}) and a complex expression defined as the list with the value of the local variable \texttt{day} and the application of the function \texttt{f} to the value of the input pipe (i.e., \texttt{[get(day,Day),apply(f(X),[In\_Arg])])}; in the function's application the Prolog's variable \emph{X} gets bounded to the current value of  \emph{In\_Arg}. The definition of \texttt{f} is given in Listing \ref{list:userfunc}. As can be observed, the value of function \texttt{f} is \emph{ok} or \emph{not ok} depending on whether the value of the input pipe is the same or different from the current value of the local variable \texttt{day}.

Each arc of \texttt{is} illustrates also a particular kind of action: \texttt{screen('Good Bye')} is a speech act that renders the expression \emph{Good Bye} when the \texttt{finish} expectation is met. The predicate \texttt{screen} is defined as a SitLog's basic action and has an associated algorithm that is executed by IOCA when it is interpreted, when its argument is rendered through speech (i.e., the robot says 'Good Bye'). The second edge illustrates a composite action: the list
\texttt{[date(get(day,Y)),next\_date(set(day,X))]}, where \texttt{date} and \texttt{next\_date} are user defined predicates --as opposed to SitLog's basic actions-- and \texttt{get} and \texttt{set} are SitLog's operators that consult and set the local variable \texttt{day}. When the corresponding expectation is met, these operators are executed and the action is grounded as the list of the two predicates with their corresponding values, and is available for inspection in the history of the task, as explained below. Finally, the action in the third edge illustrates the application of the function \texttt{g} that consults the last grounded edge traversed in the history of the task, and the action's value is the specification of such transition; the definition of \texttt{g} is given in Listing \ref{list:userfunc}.

%%%%%%%%%%%%%%%%%%%%%%
%% Listing 4
%%%%%%%%%%%%%%%%%%%%%%
\begin{listing}[tb]
\begin{Verbatim}[fontfamily=courier,fontsize=\footnotesize,frame=single,framesep=2mm]
% Second Dialogue Model
diag_mod(wait,
 % Second argument: List of Situations
 [% First situation
  [id ==> is,
   type ==> speech,
   in_arg ==> In_Arg,
   arcs ==>
        [
         In_Arg:[inc(g_count_fs1, G1)] => fs1,
         loop:[inc(g_count_fs2, G2)] => fs2
	    ]
  ],
  % Final Situation 1
  [id ==> fs1, type ==> final],
  % Final Situation 2
  [id ==> fs2, type ==> final]
 ], % End List of Situations
 % Third argument: local variables (empty)
 [ ]
). % End DM (wait)
\end{Verbatim}
\caption{SitLog's Specification of the DM {\em wait}.}
\vspace{-5pt}
\label{list:dummy_wait}
\end{listing}
%%% End listing

%%%%%%%%%%%%%%%%%%%%%%
%% Listing 5
%%%%%%%%%%%%%%%%%%%%%%
\begin{listing}[tb]
\begin{Verbatim}[fontfamily=courier, frame=single, framesep=2mm]
%Example of user functions structure
f(X) :- var_op(get(day, Day)),
        (X == Day -> Y = ok |
         otherwise -> Y = 'not ok'),
        % Assign function value
        assign_func_value(Y).
% Example of function consulting the history
g(_) :- get_history(History),
        get_last_transition(History,Last),
        % Assign function value
        assign_func_value(Last).
% Example of next state selection function
h(X, Y) :- (X == Y -> Next_Sit = is |
            otherwise -> Next_Sit = rs),
           % Assign function value
           assign_func_value(Next_Sit).
\end{Verbatim}
\caption{User functions of the dummy application.}
\vspace{-5pt}
\label{list:userfunc}
\end{listing}
%%% End listing

The first two arcs illustrate the concrete specification of next situations --\texttt{fs} and \texttt{is} respectively-- and  the third one shows the functional specification of the next situation through the function \texttt{h}, whose arguments are the current input pipe value and the current value of the local variable \texttt{day}, the latter value is conveyed in the Prolog's variable \texttt{Day}. The definition of \texttt{h} is given in Listing \ref{list:userfunc} too.

User functions are defined as standard Prolog programs that can access the current SitLog's environment (i.e., the local and global variables) as well as the history of the task through SitLog's operators, whose execution is finished with the special predicate \texttt{assign\_func\_value(Next\_Sit)}, as can be seen in Listing \ref{list:userfunc}. 

The conceptual and deliberative resources used on the demand during the interpretation of situations are defined as user functions. There is a set of user functions to retrieve information and update the content and structure of the knowledge-base service, and also to diagnose, make decisions and induce and execute plans during the interpretation of situations and dialogue models.

The second situation \texttt{rs} is of type \texttt{recursive}. It also has a local program that increments the local variable  \texttt{count\_rec} each time the corresponding embedded DM   \emph{wait} is called upon. This DM is specified by the attribute \texttt{embedded\_dm}. Recursive situations consists of control information, and the \texttt{arcs} attribute includes only the exit edges, that depend on the final state in which the embedded DM terminates; the expectation of each arch is an atom with the name of the corresponding final situation of the embedded DM (i.e., \texttt{fs1} and \texttt{fs2}). The corresponding \texttt{screen} actions render a messages through speech, as previously explained.

Final situations do not have exit edges and are specified simply by their ids and the designated type \texttt{final}. When a situation of this type is reached in the main DM the whole SitLog's program is terminated; otherwise, when a final situation of an embedded DM is reached, the control is passed back to the embedding DM, which is popped up from the DM's stack.

Finally, the third argument of the \emph{main} DM is the list of its local variables \texttt{[day ==> monday, count\_init ==> 0, count\_rec ==> 0]}. As was mentioned these variables are only visible within  \emph{main} and are outside the scope of \emph{wait}. Hence, in the present environment DMs can see their local varibales and the global variables defined for the whole SitLog application, but local variables are not seen in embedded DMs. This locality principle has also proved to be very helpful for the definition of complex applications.

The definition of the embedded DM \emph{wait} proceeds along similar lines. The initial situation \texttt{is} is of type \texttt{speech} too. It defines two arcs, the expectation of the first is the value of the input pipe and that of the second arc is the atom \texttt{loop}. If the speech interpreter matches the input pipe the global variable \texttt{g\_count\_fs1} is incremented, the final state \texttt{fs1} is reached and the execution of the \emph{wait} DM is terminated. Otherwise, the external input turns out to be unified with tha atom \texttt{loop}. When this latter path is selected the global variable \texttt{g\_count\_fs2} is incremented, the final state \texttt{fs2} is reached and the execution of the \emph{wait} DM is terminated. Finally, the list of local variables of \emph{wait} is empty, and this DM can only see global variables. Noticeably, since the \texttt{out\_arg} attribute is not set in the current DM, the input pipe of the \emph{main} DM propagates all the way back to the reentry point of the situation \texttt{rs}, that invokes the embedded DM \texttt{wait}.

We conclude the presentation of this dummy program with the history of an actual task, which is illustrated in Listing \ref{list:history}. The reader is invited to trace the program and the expectations that were met at each situation, with their corresponding next situations. The story of the whole task is provided by the SitLog interpreter when its execution is finished. The full Prolog's code of the SitLog's interpreter is available as a GitHub repository at \url{https://github.com/SitLog/source_code}. 

%%%%%%%%%%%%%%%%%%%%%%
%%%%%%%%%%%%%%%%%%%%%%

\begin{listing}[tb]
\begin{Verbatim}[fontfamily=courier,frame=single, framesep=2mm]
main: (is,[day(tuesday)]:
      [date(monday),next_date(tuesday)])
main: (is,[tuesday,'not ok']:
      ([day(tuesday)]:
      [date(monday),next_date(tuesday)]))
       [wait: (is,loop:[1])
        wait: (fs2,empty:empty)]
main: (rs,fs2:
      screen(Cont. recursive sit))
       [wait: (is,tuesday:[1])
        wait: (fs1,empty:empty)]
main: (rs,fs1:
      screen(Back to initial sit))
main: (is,[day(monday)]:
      [date(tuesday),next_date(monday)])
main: (is,[monday,ok]:
      ([day(monday)]:
      [date(tuesday),next_date(monday)]))
main: (is,finish:screen(Good Bye))
main: (fs,empty:empty)

Out Arg: monday
Out Global Vars: [g_count_fs1==>1,
                  g_count_fs2==>1]
\end{Verbatim}
\caption{History of a session.}
\vspace{-5pt}
\label{list:history}
\end{listing}

\section{Conceptual Inference Scenario}
\label{sec:preferences-scenario}
The human experience and the robot's performance is improved during the execution of a task by the robot knowing the things the user likes, social patterns, healthy guidelines, etc. Such aspects can be expressed in the KB as preferences, or conditional defaults, that helps resolving conflicting situations arising  from incompatible conclusions. Let $CD$ be the set of conditional defaults:
\[ 
\begin{array}{lcl}
    \{[Ant_{11},Ant_{12},\ldots, Ant_{1i}  & \rightarrow & Con_1,\ W_1],\\ 
    
    \;\;[Ant_{21},Ant_{22},\ldots, Ant_{2j} & \rightarrow & Con_2,\ W_2],\\ 
    
    & \vdots & \\ 
    
    \;\;[Ant_{n1},Ant_{n2},\ldots, Ant_{nm} & \rightarrow & Con_m,\ W_m]\},
\end{array}
\]
where each element in $CD$ is of the form $[Antecedents\rightarrow Consequent, Weight]$, with $Antecedents$ a list such that the $Consequent$ appended to $Antecedents$ compose a list of either properties or relations alone. Furthermore, assume that at some point in the execution of the task all antecedents are satisfied for some (more than one) conditional defaults; therefore, the corresponding consequents are also satisfied, which may cause a problem since they might be representing incompatible conclusions. This problem is solved by the Principle of Specificity applied to the weight of the conditional defaults; thus only one consequent will be considered, the one whose associated weight is the lowest. The structure of the conceptual inference scenario can be broken up in three parts: (i) {\it retrieving user preferences and getting the order}, (ii) {\it fetching and delivering items} and (iii) {\it updating the KB and applying abductive reasoning}, each one will be explained in detail next.

\subsection*{Retrieving user preferences and getting the order}
The preferences of the human user, and all relevant information for a successful interaction, should be present in the KB. This is likely to be a dynamic process since user preferences, healthy guidelines, designated home locations, items in the shelves, and so on may greatly vary from time to time. One way to keep the KB updated, probably the optimal way, is to directly querying the user. For example, if there are $k$ different new  drinks, the robot proceeds by repeatedly taking two drinks at a time and asking the user to choose the preferred one that he or she would like to be served, thus the total number of queries is $O(k
^2)$ to get the appropriate weight of all drinks with respect to the user preference. 

Once the preferences are known to the robot, it can make use of them in the course of the daily routine to reason about the state of its surroundings, the conduct of the user and the speech acts it is faced with. In the present scenario the robot offers its assistant to the user, who replies by asking for comestible objects $o_1,\ldots,o_l$ to be fetched to him. For each object $o_i$, the robot examines whether $o_i$ is the preferred object to be served among the individuals of its class, $C_{o_i}$. If so, $o_i$ is added to the list $L_{final}$ of final objects to be delivered. Otherwise, the preferred object to be served $o_{pref}$ of the class under consideration $C_{o_i}$ is obtained, and the user is queried to choose between the object he originally asked for $o_i$ and the preferred one $o_{pref}$. The user's choice is added to $L_{final}$. 

It can be noticed that getting the preferred member of a class $C$ is an important operation. Recall that preferences are conceived as conditional defaults bound to their weight, so the lower the weight the higher its preference. The steps involved in finding the preferred value of a property or relation defined in the class $C$ are:
\begin{enumerate}
    \item Retrieve from the KB the list of conditional defaults defined within $C$ and its ancestor classes.
    \item Let $\ell_{sorted}$ be the result of sorting in increasing order the list generated in the previous step. The key to sort this list is the weight value defined within the conditional defaults.
    \item For each conditional default in $\ell_{sorted}$ verify if its antecedents are satisfied. In that case, keep its consequent, which is a property or relation. Otherwise, dismiss the conditional default. Then, delete from left to right consequents that define a property or relation more that once, preserving the first occurrence. Let $\ell_{del}$ be the list that is obtained after this deletion.
    \item In $\ell_{del}$ find the property or relation of interest whose value is the desired output.
\end{enumerate}

 For the situation described above, the preferred object is sought as the argument of the property {\it to serve}, ocurring in the consequent of the conditional defaults present in the class $C_{o_i}$.

Interestingly, the robot can adequately deal with user commands that are underspecified, i.e., commands asking for an individual object but missing specific information that uniquely identifies it, providing instead general information. For instance, {\it bring me something to drink}. The robot deals with this kind of commands by taking the preferred individual of the class being asked.

Therefore, at the end of the speech act, whether the user requests objects by name or by giving general information, the robot is able to formulate the final list $L_{final}$ of objects to be fetched to the user.

\subsection*{Fetching and delivering items}
For each $o_i\in L_{final}$, the robot queries the KB to retrieve the list $[loc_1,\ldots,loc_n]$ of preferred locations where $o_i$ is likely to be found, such that $loc_1$ is the most preferred location and $loc_n$ is the least preferred location. Furthermore, $[loc_1,\ldots,loc_n]$ is a permutation of the locations $l_1,\ldots,l_n$ in $L$ (see the settings of the conceptual inference in Section \ref{sec:conceptual-inference}).

Obtaining the list of preferred locations of an object is an operation closely related to the operation outlined above that finds the preferred member of a class. The steps are:
\begin{description}
    \item [Step 1] retrieves not only the conditional defaults of the object's class and its ancestors, but also the conditional defaults of the object itself.
    \item [Step 2] is the same of that to find the preferred member of a class.
    \item [Step 3] keeps the consequent of conditional defaults whose antecedents are satisfied but does not delete any of them, although a property or relation may be defined multiple times.
    \item [Step 4] subtracts in order all the values of the property or relation of interest, producing thus the desired list.
\end{description}
For the list of preferred locations needed by the robot, the property of interest in step 4 is the location, or $loc$ as it is defined in the KB.

Next, the robot visits the shelves at the locations $[loc_1,\ldots,loc_n]$ in their order of appearance, searching for the object $o_i$ in each of them. If the object $o_i$ is found at $loc_j$, the robot takes it, and repeats now the process for the object $o_{i+1}$ in $L_{final}$. If $o_i$ is not found at $loc_j$, the robot searches for it in the shelf located at $loc_{j+1}$. When an error arises taking an object, moving to a new location, or realizing that the object is not found after visiting all shelves, then a recovery protocol can be invoked or the daily life inference cycle triggered. 

At this point two important observations have to be made:
\begin{enumerate}
    \item As a side effect of searching for an object, the properties of other objects in the robot's KB may change. Suppose that the robot makes an observation in the shelf located at $loc_j$ trying to find object $o_i$. Regardless of recognizing $o_i$, the robot may have seen a set $O=\{o_{i1},\ldots,o_{i\bar{n}}\}$ of objects. Hence, the robot is now aware of the precise shelf where such objects are placed. So, the property {\it last seen} for these objects is assigned to $loc_j$ in the KB. Therefore, for any $obj\in O$ the first element of its list of preferred locations has to be $loc_j$. The KB works in this way since a conditional default for the object $obj$ is defined with antecedent {\it last seen}, consequent $loc$ and weight 1, as it is seen in Figure \ref{diag:kb}.
    \item When the robot takes two objects, using its two hands, or when it is holding one object and there are no more left to take, the objects must be delivered to the user. But first, the robot needs to determine the room where the user may have gone based on his or her preferences and properties. The preferred room is retrieved from the KB by an operation that can be derived from the steps explained above. In the current scenario, two conditional defaults for the individual {\it user} have been defined whose consequent is the property {\it found in} that indicates the room where the user is located, and whose antecedents are conditions that cause him or her to go to one room or to another depending on the user's mood or physical state, as shown also in Figure \ref{diag:kb}. After this delivery, the robot examines the list $L_{final}$ to know whether there are more objects to be fetched or not.
\end{enumerate}

Noticeably, the inference mechanism on conditional defaults handles chained implications, so it is plausible to have a conditional default such that its antecedents may be satisfied by the consequents of other conditional defaults. Let $Prop$ be the list of known closed propositions (properties or relations) and $L_{CD}$ be the list $[Ant\rightarrow Consequent\ |\  More_{CD}]$ of conditional defaults of a class or individual, such that the conditional defaults in $L_{CD}$ are sorted in increasing order with respect to their weight, value that is omitted in $L_{CD}$. The mechanism proceeds recursively over the head element $Ant\rightarrow Consequent$ of $L_{CD}$ according to the cases:
\begin{itemize}
    \item $Ant$ is a single property or relation. Examine all valid pattern matching situations for $Ant\rightarrow Consequent$ as follows: (a) Since $Ant$ and $Consequent$ are both properties or both relations, their pattern is $\alpha=>\beta$; nonetheless, $\beta$ may be a variable, and the pattern becomes $\alpha=>'\!\!-'$. (b) A property may be a single label with no associated value. (c) $Ant$ may be absent, which is represented as $'\!-'\!\rightarrow Consequent$, indicating that the consequent of the conditional default is always satisfied.
    \begin{enumerate}
        \item Now, execute a backward analysis checking whether $Ant$ is already part of $Prop$, if so add the corresponding $Consequent$ to $Prop$. 
        \item Otherwise, execute a forward analysis verifying if $Ant$ occurs as the consequent of a conditional default in $More_{CD}$; in that case apply the current analysis to the list $More_{CD}$, whose output is a temporary set of new closed propositions $Tem_{Prop}$, such that $Consequent$ is added to $Prop$ whenever $Ant$ is part of $Tem_{Prop}$.
    \end{enumerate} 
    Remarkably, matching $Ant$ with an element of $Prop$ or $Tem_{Prop}$ instantiates the variable that $Ant $ might have. Since the variable in the antecedent is bound in the consequent, instantiating the variable in $Ant$ provides a value for the variable in $Consequent$.
    \item $Ant$ is a list of properties or relations. Check whether the elements in the list $Ant$ are part of $Prop$ or $Tem_{Prop}$, as explained above. If that is true for all elements in $Ant$, then add $Consequent$ to $Prop$.
\end{itemize}

Finally, the desired property or relation is searched for in the resulting list $Prop$, and its first occurrence is output.

This mechanism is illustrated in our scenario as the user is assumed to be home after a bad day at work and to have requested comestible objects. Therefore, the first conditional default for the individual $user$ implies that he or she is also tired, this consequent is chained to the next implication since being tired and back from work implies that the user is found at the living room; but having requested comestible objects implies that he or she is in a different room. The conditional default concluding that the user is in the living room has lower weight, so that room is the preferred one where the user may be found.
    
\subsection*{Updating the KB and applying abductive reasoning}
Once the robot reaches the user hands over the objects it is carrying. At this point, the robot knows the location from where such objects were taken. For each delivered object $o_{del}$, if there is an inconsistency between the preferred location to find $o_{del}$ as stated in the KB, $loc_{pref}$, and the actual location where $o_{del}$ was taken, $loc_{act}$, then the robot informs the user of this situation and asks him to choose the preferred location between $loc_{pref}$ and $loc_{act}$. The  data for $o_{del}$ in KB is updated when the user picks $loc_{act}$ over $loc_{pref}$.

After delivering all requested objects, the robot examines the location of other objects that it may have seen during the execution of the whole task. As described above, for the observed objects the property {\it last seen} in the KB is updated with the location where they were last seen. Let $O'=\{o'_1,\ldots, o'_t\}$ be the set of non-requested objects seen while executing the task. For each $o'_i\in O'$:

\begin{itemize}
    \item The robot retrieves from the KB the value of the property {\it last seen} for $o'_i$, denoted $loc_{last}$, and the list $[loc_1,loc_2,\ldots,loc_{n}]$ of preferred locations where $o'_i$ is likely to be found. Now, $loc_{last}$ and $[loc_1,loc_2,\ldots,loc_{n}]$ are examined to determine any inconsistency. In fact, $loc_{last}=loc_1$ since the location where $o'_i$ was last seen has the highest preference; $loc_2$ is next as the predefined location where $o'_i$ is more likely to be found. If $loc_{last}\neq loc_2$, then a problem is detected having seen $o'_i$ not in its predefined location. Thus, the robot defines the property {\it misplaced} for $o'_i$ in the KB. 

    \item The abductive reasoner is triggered to find out a possible explanation for the misplacement of $o'_i$. This reasoner takes the lists $L_{CD}$ and $Prop$, as defined previously, and recursively examines the pattern for each conditional default $Ant\rightarrow Consequent$ in $L_{CD}$ similarly to the inference mechanism described above, but $Consequent$ is checked to belong to $Prop$ instead of $Ant$. If this check turns out to be true, then the pair $Consequent:Ant$ is added to the list of explanations. After all conditional defaults are analyzed, the list of explanations is trimmed by keeping the first occurrence of a pair with a given $Consequent$ and removing all others. This respects the order of preference since the explanation drawn from the conditional default with lowest weight is kept.

    \item The application of the abductive reasoner on the current scenario reveals, by the conditional defaults defined in the class $object$, that $o'_i$ is misplaced because it was moved by the user's child or by the user's partner. The weight associated to each conditional default is considered to conclude that $o'_i$ is misplaced because the user's child moved it to $loc_{last}$. Finally, if allowed by the user, the robot goes to $loc_{last}$, takes $o'_i$ and places it in $loc_2$. After this, the robot is finished examining $o'_i$.
\end{itemize}

\bibliographystyle{plain}
\bibliography{bibliografia}

\begin{thebibliography}{10}

\bibitem{becker2011pr2}
Jan Becker, Christian Bersch, Dejan Pangercic, Benjamin Pitzer, Thomas
  R{\"u}hr, Bharath Sankaran, J{\"u}rgen Sturm, Cyrill Stachniss, Michael
  Beetz, and Wolfram Burgard.
\newblock The pr2 workshop-mobile manipulation of kitchen containers.
\newblock In {\em IROS workshop on results, challenges and lessons learned in
  advancing robots with a common platform}, volume 120, 2011.

\bibitem{smach}
Jonathan Bohren, Radu~Bogdan Rusu, E.~Gil Jones, Eitan Marder-Eppstein,
  Caroline Pantofaru, Melonee Wise, Lorenz M\"osenlechner, Wim Meeussen, and
  Stefan Holzer.
\newblock Towards autonomous robotic butlers: Lessons learned with the pr2.
\newblock In {\em Proc. of the International Conference on Robotics and
  Automation}, pages 5568--5575, 2011.

\bibitem{Chen2014}
Kai Chen, Dongcai Lu, Yingfeng Chen, Keke Tang, Ningyang Wang, and Xiaoping
  Chen.
\newblock The intelligent techniques in robot kejia -- the champion of
  robocup@home 2014.
\newblock In Reinaldo A.~C. Bianchi, H.~Levent Akin, Subramanian Ramamoorthy,
  and Komei Sugiura, editors, {\em RoboCup 2014: Robot World Cup XVIII}, pages
  130--141, 2015.

\bibitem{Chen2010}
Xiaoping Chen, Jianmin Ji, Jiehui Jiang, Guoqiang Jin, Feng Wang, and Jiongkun
  Xie.
\newblock Developing high-level cognitive functions for service robots.
\newblock In {\em Proceedings of the International Conference on Autonomous
  Agents and Multiagent Systems}, volume~1, pages 989--996, 2010.

\bibitem{Chen2013}
Xiaoping Chen, Jianmin Ji, Zhiqiang Sui, and Jiongkun Xie.
\newblock {Handling open knowledge for service robots}.
\newblock In {\em International Joint Conference on Artificial Intelligence},
  pages 2459--2465, 2013.

\bibitem{Chen2015}
Yingfeng Chen, Feng Wu, Wei Shuai, Ningyang Wang, Rongya Chen, and Xiaoping
  Chen.
\newblock Kejia robot--an attractive shopping mall guider.
\newblock In Adriana Tapus, Elisabeth Andr{\'e}, Jean-Claude Martin,
  Fran{\c{c}}ois Ferland, and Mehdi Ammi, editors, {\em Social Robotics}, pages
  145--154, 2015.

\bibitem{trex}
Sachin Chitta, E.~Gil Jones, Matei Ciocarlie, and Kaijen Hsiao.
\newblock Mobile manipulation in unstructured environments: Perception,
  planning, and execution.
\newblock {\em {IEEE} Robotics and Automation Magazine}, 19(2):58--71, 2012.

\bibitem{moped}
Alvaro Collet, Manuel Martinez, and Siddhartha~S. Srinivasa.
\newblock The {MOPED} framework: Object recognition and pose estimation for
  manipulation.
\newblock {\em The International Journal of Robotics Research}, 30:1284--1306,
  2011.

\bibitem{Durham2008}
J.W. Durham and F.~Bullo.
\newblock {Smooth Nearness-Diagram Navigation}.
\newblock In {\em {Intelligent Robots and Systems, 2008. IROS 2008. IEEE/RSJ
  International Conference on}}, pages 690--695, Sept 2008.

\bibitem{Espinace2013}
Pablo Espinace, Thomas Kollar, Nicholas Roy, and Alvaro Soto.
\newblock Indoor scene recognition by a mobile robot through adaptive object
  detection.
\newblock {\em Robotics and Autonomous Systems}, 61(9):932---947, 2013.

\bibitem{fan2014}
Zhengjie Fan, Elisa Tosello, Michele Palmia, and Enrico Pagello.
\newblock Applying semantic web technologies to multi-robot coordination.
\newblock In {\em Proceedings of the International Conference Intelligent
  Autonomous Systems}, 2014.

\bibitem{pomdp_nav}
Amalia Foka and Panos Trahanias.
\newblock Real-time hierarchical pomdps for autonomous robot navigation.
\newblock {\em Robotics and Autonomous Systems}, 55(7):561--571, 2007.

\bibitem{readylog}
C.~Fritz.
\newblock {Integrating Decision-Theoretic Planning and Programming for Robot
  Control in Highly Dynamic Domains}.
\newblock Master's thesis, RWTH Aachen University, Knowledge-based Systems
  Group, Aachen, Germany, 2003.

\bibitem{galindo2008}
Cipriano Galindo, Juan-Antonio Fern\'andez-Madrigal, Javier Gonz\'alez, and
  Alessandro Saffiotti.
\newblock Robot task planning using semantic maps.
\newblock {\em Robotics and Autonomous Systems}, 56(11):955--966, 2008.

\bibitem{Grisetti2007}
G.~Grisetti, C.~Stachniss, and W.~Burgard.
\newblock {Improved Techniques for Grid Mapping With Rao-Blackwellized Particle
  Filters}.
\newblock {\em Robotics, IEEE Transactions on}, 23(1):34--46, Feb 2007.

\bibitem{golex}
Dirk H\"{a}hnel, Wolfram Burgard, and Gerhard Lakemeyer.
\newblock Golex--bridging the gap between logic (golog) and a real robot.
\newblock In Otthein Herzog and Andreas Günter, editors, {\em Advances in
  Artificial Intelligence}, volume 1504, pages 165--176. Springer Berlin /
  Heidelberg, 1998.

\bibitem{pomdp_manipulation}
K.~Hsiao, L.~P. Kaelbling, and T.~Lozano-Perez.
\newblock Grasping pomdps.
\newblock In {\em Proceedings of the IEEE International Conference on Robotics
  and Automation}, pages 4685--4692, 2007.

\bibitem{jakob_asp}
S.~{Jakob}, S.~{Opfer}, A.~{Jahl}, H.~{Baraki}, and K.~{Geihs}.
\newblock Handling semantic inconsistencies in commonsense knowledge for
  autonomous service robots.
\newblock In {\em IEEE International Conference on Semantic Computing}, pages
  136--140, 2020.

\bibitem{Karg2012}
M.~Karg and A.~Kirsch.
\newblock Acquisition and use of transferable, spatio-temporal plan
  representations for human-robot interaction.
\newblock In {\em Proceedings of the IEEE/RSJ International Conference on
  Intelligent Robots and Systems}, pages 5220--5226, 2012.

\bibitem{Lim2011}
Gi~Hyun Lim, Il~Hong Suh, and Hyowon Suh.
\newblock Ontology-based unified robot knowledge for service robots in indoor
  environments.
\newblock {\em IEEE Transactions on Systems, Man, and Cybernetics Part
  A:Systems and Humans}, 41(3):492--509, 2011.

\bibitem{xabsl}
Martin Lotzsch, Max Risler, and Matthias J{\"u}ngel.
\newblock {XABSL} - {A} pragmatic approach to behavior engineering.
\newblock In {\em Proceedings of {IEEE/RSJ} International Conference of
  Intelligent Robots and Systems}, pages 5124--5129, 2006.

\bibitem{Marr:1982:VCI:1095712}
David Marr.
\newblock {\em Vision: A Computational Investigation into the Human
  Representation and Processing of Visual Information}.
\newblock Henry Holt and Co., Inc., New York, NY, USA, 1982.

\bibitem{opfer_asp}
Stephan Opfer, Stefan Jakob, and Kurt Geihs.
\newblock Teaching commonsense and dynamic knowledge to service robots.
\newblock In Miguel~A. Salichs, Shuzhi~Sam Ge, Emilia~Ivanova Barakova,
  John-John Cabibihan, Alan~R. Wagner, {\'A}lvaro Castro-Gonz{\'a}lez, and
  Hongsheng He, editors, {\em Social Robotics}, pages 645--654, 2019.

\bibitem{Pangercic}
D.~Pangercic, M.~Tenorth, D.~Jain, and M.~Beetz.
\newblock Combining perception and knowledge processing for everyday
  manipulation.
\newblock In {\em IEEE/RSJ International Conference on Intelligent Robots and
  Systems}, pages 1065--1071, 2010.

\bibitem{dime-damls}
Luis Pineda, Varinia Estrada, Sergio Coria, and James Allen.
\newblock The obligations and common ground structure of practical dialogues.
\newblock {\em Inteligencia Artificial, Revista Iberoamericana de Inteligencia
  Artificial}, 11:9--17, 12 2007.

\bibitem{ioca-2011}
Luis~A. Pineda, Ivan Meza, Hector Aviles, Carlos Gershenson, Caleb Rascon,
  Monserrat Alvarado, and Lisset Salinas.
\newblock Ioca: Interaction-oriented cognitive architecture.
\newblock {\em Research in Computing Science}, 54:273--284, 2011.

\bibitem{service-robot-2014}
Luis~A. Pineda, Arturo Rodriguez, Gibran Fuentes, Caleb Rascon, and Ivan~V.
  Meza.
\newblock Concept and functional structure of a service robot.
\newblock {\em International Journal of Advanced Robotic Systems}, pages 1--15,
  2013.

\bibitem{Non-Monotonic-KB-2017}
Luis~A. Pineda, Arturo Rodr\'iguez, Gibran Fuentes, Caleb Rasc\'on, and Ivan~V.
  Meza.
\newblock A light non-monotonic knowledge-base for service robots.
\newblock {\em Intel Serv Robotics}, 10:159--171, 2017.

\bibitem{sitlog-2013}
Luis~A. Pineda, Lisset Salinas, Ivan Meza, Caleb Rascon, and Gibran Fuentes.
\newblock $sitlog$: A programming language for service robot tasks.
\newblock {\em International Journal of Advanced Robotic Systems}, pages 1--12,
  2013.

\bibitem{reiter-1980}
R.~Reiter.
\newblock A logic for default reasoning.
\newblock {\em Artificial Intelligence}, 13:81--132, 1980.

\bibitem{Schiffer2012}
Stefan Schiffer, Alexander Ferrein, and Gerhard Lakemeyer.
\newblock Caesar: an intelligent domestic service robot.
\newblock {\em Intelligent Service Robotics}, 5(4):259--273, 2012.

\bibitem{schiffer_golog}
Stefan Schiffer, Alexander Ferrein, and Gerhard Lakemeyer.
\newblock Reasoning with qualitative positional information for domestic
  domains in the situation calculus.
\newblock {\em Journal of Intelligent and Robotic Systems}, 66(1--2):273--300,
  2012b.

\bibitem{tdl}
Reid Simmons and David Apfelbaum.
\newblock A task description language for robot control.
\newblock In {\em Proc. of the Conference on Intelligent Robots and Systems},
  1998.

\bibitem{herb2.0}
Siddhartha Srinivasa, Dmitry Berenson, Maya Cakmak, Alvaro {Collet Romea},
  Mehmet Dogar, Anca Dragan, Ross~Alan Knepper, Tim~D Niemueller, Kyle
  Strabala, J~Michael Vandeweghe, and Julius Ziegler.
\newblock Herb 2.0: Lessons learned from developing a mobile manipulator for
  the home.
\newblock {\em Proceedings of the {IEEE}}, 100(8):1--19, 2012.

\bibitem{knowrob-map}
M.~Tenorth, L.~Kunze, D.~Jain, and M.~Beetz.
\newblock Knowrob-map - knowledge-linked semantic object maps.
\newblock In {\em IEEE-RAS International Conference on Humanoid Robots}, pages
  430--435, 2010.

\bibitem{tenorth2013knowrob}
Moritz Tenorth and Michael Beetz.
\newblock Knowrob: A knowledge processing infrastructure for cognition-enabled
  robots.
\newblock {\em The International Journal of Robotics Research}, 32(5):566--590,
  2013.

\bibitem{Tenorth2017}
Moritz Tenorth and Michael Beetz.
\newblock Representations for robot knowledge in the knowrob framework.
\newblock {\em Artificial Intelligence}, 247:151--169, 2017.

\bibitem{Preferences-SR-2018}
Ivan Torres, No\'e Hern\'andez, Arturo Rodr\'iguez, Gibr\'an Fuentes, and
  Luis~A. Pineda.
\newblock Reasoning with preferences in service robots.
\newblock {\em Journal of Intelligent \& Fuzzy Systems}, 36(5):5105--5114,
  2019.

\bibitem{xrobots}
Steve Tousignant, Eric Van~Wyk, and Maria Gini.
\newblock An overview of {XRobots}: A hierarchical state machine-based
  language.
\newblock In {\em Proc. of the Workshop on Software Development and Integration
  in Robotics}, 2011.

\bibitem{Zhang2012a}
Shiqi Zhang, Mohan Sridharan, and Forrest {Sheng Bao}.
\newblock {ASP+POMDP: Integrating non-monotonic logic programming and
  probabilistic planning on robots}.
\newblock In {\em Proceedings of the IEEE International Conference on
  Development and Learning and Epigenetic Robotics}, 2012.

\end{thebibliography}

\end{document}